\documentclass{article}

\usepackage[main, final]{neurips_2026}
\usepackage{graphicx}
\usepackage{amsmath}
\usepackage{algorithm}
\usepackage{algpseudocode}
\usepackage{wrapfig}
\usepackage{caption}

\usepackage[utf8]{inputenc}
\usepackage[T1]{fontenc}
\usepackage{hyperref}
\usepackage{url}
\usepackage{booktabs}
\usepackage{amsfonts}
\usepackage{nicefrac}
\usepackage{microtype}
\usepackage{xcolor}
\usepackage{multirow}

\newcommand{\daeun}[1]{#1}

\title{Disentangling Channel Semantics in \\Vision Transformers via Token Decorrelation and Composition-Aware Modulation}

\author{
  Daeun Kim$^{1}$,
  Hyesong Choi$^{2}$,
  Hyejin Park$^{1,\dagger}$,
  Dongbo Min$^{1,\dagger}$ \\
  $^{1}$Division of Artificial Intelligence \& Software,
  Ewha Womans University \\
  $^{2}$Soongsil University \\
  \texttt{\{daeunkim00,clrara,dbmin\}@ewha.ac.kr} \\
  \texttt{hyesong@ssu.ac.kr}
}

\begin{document}

\maketitle
{\renewcommand{\thefootnote}{\fnsymbol{footnote}}%
 \footnotetext[2]{Corresponding authors.}}

\begin{abstract}
Multi-channel images encode heterogeneous channel semantics aligned over a shared spatial structure, which requires models to jointly capture globally shared structure and channel-specific variations.
Recent multi-channel Vision Transformers attempt to address this challenge by augmenting the \texttt{[CLS]} token with memory tokens to organize multi-channel representations under varying channel compositions.
However, we find that these context tokens collapse onto a few dominant channels under high inter-channel redundancy, producing redundant representations and a biased global summary that fails to reliably support composition-aware channel interpretation.
We propose \textbf{MuCa-ViT (Multi-Channel Composition-Aware Vision Transformer)}, a unified framework that addresses these limitations through two coupled mechanisms.
\textbf{Factorial Token Learning (FTL)} enforces token-level decorrelation among context tokens, encouraging the \texttt{[CLS]} token to capture globally shared semantics while memory tokens preserve complementary channel-specific cues.
\textbf{Channel Composition-Aware Modulation (CAM)} uses the FTL-refined \texttt{[CLS]} as a semantic anchor to generate sample-wise modulation signals, which enable composition-aware interpretation of channels within a unified backbone rather than relying on static channel embeddings.
Experiments on JUMP-CP, CHAMMI, and So2Sat show that MuCa-ViT outperforms prior multi-channel ViTs across all three benchmarks, and the magnitude of improvement aligns with the inter-channel redundancy of each dataset ($\rho \in [0.19, 0.77]$), validating the diagnostic analysis that
motivates our design. Our code is available on \href{https://github.com/kimdanni/MuCa-ViT}{GitHub}.
\end{abstract}
\section{Introduction}

While Vision Transformers (ViTs)~\citep{dosovitskiy2020image,liu2021swin} have advanced natural image analysis through their ability to capture long-range dependencies, their extension to multi-channel imaging introduces fundamental challenges. Unlike RGB images, whose channels exhibit strong spectral correlation and jointly encode a single visual object, multi-channel images encode heterogeneous channel semantics aligned over a shared spatial structure. Effectively modeling such data requires jointly learning global representations and fine-grained, channel-specific semantic factors.
In real-world biomedical imaging and remote sensing applications, the number and type of available channels vary across samples due to acquisition protocols, sensor heterogeneity, or missing modalities, further increasing the complexity of representation learning. Consequently, multi-channel ViTs must handle inputs in which heterogeneous channel-specific information coexists on top of shared structural information, while organizing such partially observed inputs into a coherent internal representation.

Existing multi-channel ViTs typically encode channel identity using static channel embeddings while serializing channel-wise patches into a unified token sequence~\citep{pham2025cha,bourriez2024chada,Pham2024}.
To further aggregate multi-channel information, recent approaches introduce additional memory tokens alongside the \texttt{[CLS]} token~\citep{pham2025cha}. We collectively refer to these tokens as \textbf{Context Semantic Tokens (CSTs)}. However, whether globally shared and channel-specific semantics are
separately captured across these tokens, and whether channel representations
adapt to the observed channel composition, remain underexplored.

To examine these questions, we identify two co-occurring limitations in existing multi-channel ViTs. First, in the absence of any structural constraint on CSTs, both the [CLS] token and memory tokens collapse onto a small subset of dominant channels, producing highly redundant representations. This pattern arises because multi-channel inputs contain both globally
shared structure and channel-specific information.
Under high inter-channel redundancy, the model learns a shortcut by relying on a few channels with the strongest signal rather than attending to channel-specific cues. As a result, the global representation becomes biased toward dominant channels and no longer reliably reflects the full channel composition of the input. Second, existing models lack a mechanism to interpret each channel relative to the observed channel composition. The semantic role of a channel depends on which other channels are present, yet static channel embeddings remain fixed regardless of the input composition. Combined with the collapse of CSTs onto dominant channels, this absence of composition-aware interpretation leaves channel features entangled around a few dominant channels rather than separated into channel-specific factors.

To address these two limitations, we propose \textbf{MuCa-ViT} (Multi-Channel Composition-Aware Vision Transformer), a unified framework built on two design choices that have not been jointly considered in prior multi-channel ViTs. First, we introduce \textbf{Factorial Token Learning (FTL)}, which enforces explicit token-level decorrelation among context tokens to encourage auxiliary memory tokens to preserve complementary information rather than inheriting the channel preferences of the [CLS] token. This suppresses redundancy among CSTs and
yields a reliable global summary. Second, we introduce Channel \textbf{Composition-Aware Modulation (CAM)}, which complements static channel identity encoding with conditioning derived from the input channel composition, allowing the same channel to be interpreted differently depending on the observed channel subset. By using the refined [CLS] token as a semantic anchor, CAM enables composition-aware interpretation of channel features. When combined, FTL refines the conditioning signal that CAM relies on, yielding larger gains than either component alone, particularly under high inter-channel redundancy.

Our contributions are summarized as follows:

\textbf{(i) Diagnostic analysis of multi-channel ViTs.} We analyze how inter-channel redundancy promotes context token collapse onto dominant
channels in multi-channel ViTs, and how this collapse undermines composition-aware channel interpretation.

\textbf{(ii) Principled complementary design of FTL and CAM.} Motivated by our analysis, we propose MuCa-ViT, a framework with a novel complementary design that addresses collapse at two coupled levels.

\textbf{(iii) Generalizable Improvements.} We demonstrate consistent improvements across three multi-channel benchmarks and two backbone architectures, with the magnitude of each gain quantitatively tracking inter-channel redundancy as predicted by our analysis.
\section{Related Works}

{\textbf{Vision Transformers for Multi-Channel Imaging.}}
{Modeling multi-channel biomedical imaging requires handling heterogeneous channel configurations and varying channel availability. Early approaches to channel adaptation included template-mixing methods that construct channel-specific input projections from shared parameter templates~\citep{plummer2020neural,savarese2019learning}. } {With Vision Transformers (ViTs), subsequent works serialized channel-wise patches into a unified token sequence and introduced learnable channel embeddings to preserve channel identity~\citep{bao2023channel}. To improve robustness under irregular channel compositions, DiChaViT~\citep{Pham2024} incorporated diverse channel sampling and orthogonality constraints to promote channel-wise diversity.}
{More recently, pretraining-based strategies have been explored. ChA-MAEViT~\citep{pham2025cha} framed channel dropping as a masked reconstruction problem within a joint MAE-classification framework~\citep{liang2022supmae,assran2022masked} and introduced memory tokens linked to the \texttt{[CLS]} token. IC-ViT~\citep{lian2025isolated} pretrains on single-channel inputs and integrates cross-channel dependencies during fine-tuning.}
{Despite these advances, existing approaches primarily operate at the input, embedding, or supervision level. The internal token space remains implicitly structured, and the interaction between global context and channel-specific semantics within token-based architectures therefore remains underexplored.}

\textbf{Conditional Feature Modulation.}
Feature-wise conditioning adapts representations via affine transformations driven either by external signals such as language embeddings, timesteps, or class labels~\citep{perez2018film, dhariwal2021diffusion, peebles2023scalable, zhang2025diffusion, hong2025salad}, or by the model's own intermediate features~\citep{birnbaum2019temporal, brockschmidt2020gnn}. In both cases, the conditioning signal is used as given. Extending this to multi-channel ViTs requires care with the conditioning signal itself: the \texttt{[CLS]} token is a natural internal candidate, but already reflects dominant-channel bias (Sec.~\ref{sec:info_bottleneck}), and an effective signal should be interpreted jointly with the input channel configuration. This motivates calibrating the conditioning signal with respect to the channel configuration before modulation, rather than applying it on uncorrected context tokens.
\section{Problem Formulation and Analysis}
\label{sec:info_bottleneck}

\textbf{Preliminary. }
Multi-channel images consist of $C$ channels that share a common spatial structure, where each channel encodes heterogeneous semantic information. Existing ViT-based architectures typically adopt an early-fusion strategy, in which each channel $c \in \{1,\dots,C\}$ is independently patchified into $P$ tokens and all tokens are concatenated into a single sequence of length $C \times P$~\cite{bao2023channel,bourriez2024chada,Pham2024}. To preserve channel identity, a learnable channel embedding is added to each token. To summarize channel-wise information, ViT-based architectures employ a global \texttt{[CLS]} token. In practical training protocols, channel composition is often varied by sampling different subsets of channels during training, resulting in dynamically changing channel configurations across iterations.

To improve robustness under such varying channel compositions, some recent multi-channel extensions introduce additional memory tokens that interact with patch tokens within the encoder to aggregate information across channels~\cite{pham2025cha}. In this work, we collectively refer to the \texttt{[CLS]} token and memory tokens as \textbf{Context Semantic Tokens (CSTs)}, which serve as compact representations of global and channel-specific semantics.

\begin{figure}[t!]
\centering
\includegraphics[width=\linewidth]{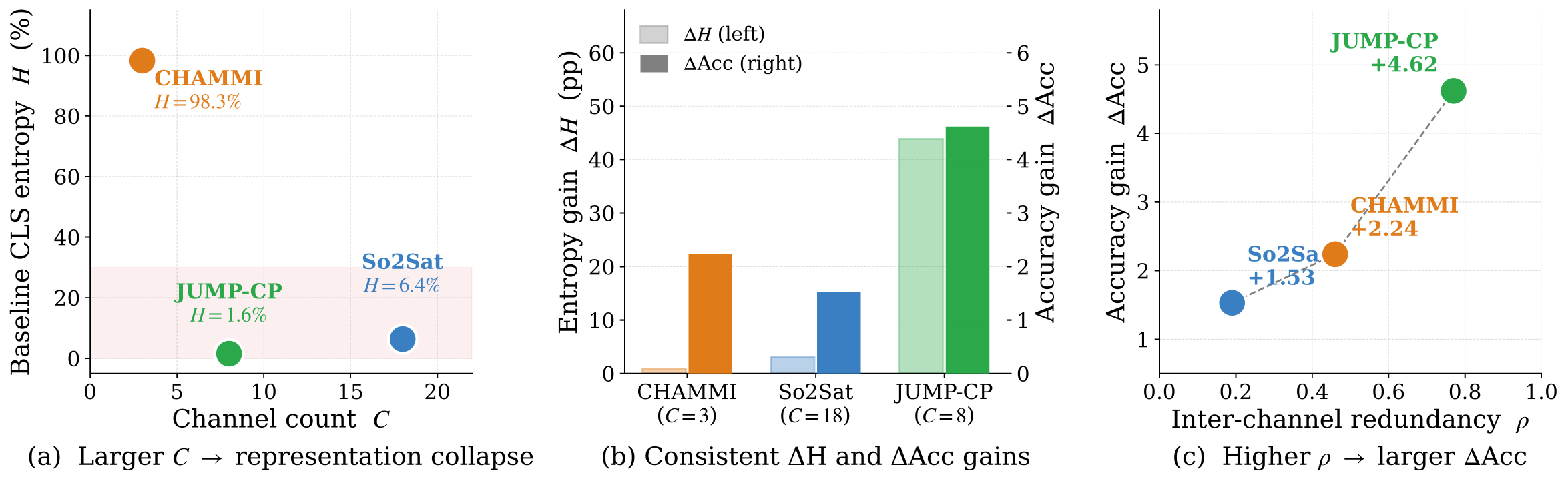}
\caption{
\textbf{Representation collapse and its mitigation across datasets.}
$\Delta$ values are computed against our reproduced ChA-MAEViT~\cite{pham2025cha}.
(a) \textbf{CLS channel coverage across datasets} ($H_{\texttt{CLS}}\%$, y-axis). Near-max on CHAMMI but drops sharply on JUMP-CP and So2Sat, indicating that CLS coverage concentrates on few channels as $C$ grows.
(b) \textbf{Consistent recovery.} Our method yields positive $\Delta H_{\texttt{CLS}}\%$ and $\Delta\mathrm{Acc}$ on all three datasets.
(c) \textbf{Larger gains in higher-redundancy regimes.}
$\Delta\mathrm{Acc}$ is larger on benchmarks with higher inter-channel redundancy $\rho$, consistent with redundancy being a major structural factor that promotes CST collapse.
}
\label{fig:channel_collapse}
\end{figure}

\noindent\textbf{Channel Dominance in CSTs.}
CSTs are intended to organize global and channel-specific information across the \texttt{[CLS]} token and auxiliary memory tokens, going beyond the limited capacity of a single global representation.
Under this design, the \texttt{[CLS]} token is expected to capture global semantics that support varying channel compositions and are forwarded to the prediction head, while memory tokens preserve complementary channel-specific cues.

To examine whether this design is actually realized in practice, or whether CSTs instead concentrate on a narrow subset of channels and undermine their intended role, we measure how broadly channels are covered within the CSTs at two complementary units, namely the \texttt{[CLS]} token and the CST set as a whole.
We analyze token interactions through similarity in the final feature space, since this representation is what is directly forwarded to the prediction head and therefore most relevant to downstream behavior. Let $\mathbf{Z} \in \mathbb{R}^{(C\times P)\times D}$ denote the final representation of patch tokens at the encoder output, where $\mathbf{Z}_{c,p}$ is the $p$-th token of channel $c$. For a query token $\mathbf{q}$ (either the \texttt{[CLS]} token or a memory token), we form a softmax distribution over the full set of $C \times P$ patch tokens based on dot-product alignment, marginalize over patches within each channel, and compute the resulting channel coverage entropy:
\begin{equation}
\small
\alpha_{c,p}(\mathbf{q}) = \frac{\exp\langle \mathbf{q},\, \mathbf{Z}_{c,p}\rangle}{\sum_{c',p'} \exp\langle \mathbf{q},\, \mathbf{Z}_{c',p'}\rangle},
\qquad
\tilde{S}_c(\mathbf{q}) = \sum_{p=1}^{P} \alpha_{c,p}(\mathbf{q}),
\qquad
H(\mathbf{q}) = -\frac{1}{\log C}\sum_{c=1}^{C} \tilde{S}_c(\mathbf{q})\,\log \tilde{S}_c(\mathbf{q}).
\label{eq:sim}
\end{equation}
The first term $\alpha_{c,p}(\mathbf{q})$ is a distribution over patches, and $\tilde{S}_c(\mathbf{q})$ aggregates patch-level attention into a channel-wise marginal representing the total attention mass that $\mathbf{q}$ places on channel $c$, where dominant channels correspond to large $\tilde{S}_c(\mathbf{q})$. The entropy $H(\mathbf{q}) \in [0,1]$ approaches $1$ when channels contribute evenly and $0$ when a few channels dominate. We instantiate $H(\mathbf{q})$ at two complementary scopes.

\noindent\emph{(i) Token-level coverage of} \texttt{[CLS]} :  $H_{\texttt{CLS}} = H(q=\texttt{[CLS]})$ which captures channel dominance from the perspective of the global summary token forwarded to the prediction head.

\noindent\emph{(ii) Set-level coverage of CSTs} :
Let $\mathcal{T} = \{q_0, q_1, \ldots, q_K\}$, where $q_0 = \texttt{[CLS]}$ and $q_1, \ldots, q_K$ are memory tokens. We aggregate the per-CST channel marginals and compute the entropy of their average:
\begin{equation}
\bar{S}_c \;=\; \frac{1}{|\mathcal{T}|}\sum_{q \in \mathcal{T}} \tilde{S}_c(q),
\qquad
H_{\text{CST}} = -\frac{1}{\log C}\sum_{c=1}^{C} \bar{S}_c\,\log \bar{S}_c.
\label{eq:cst_coverage}
\end{equation}
A high $H_{\text{CST}}$ indicates that the CSTs \emph{as a group} attend broadly across channels, while a low value indicates that they concentrate on the same dominant channels as \texttt{[CLS]}. We use $H_{\text{CST}}$ to assess whether memory tokens fulfill their intended role under missing-channel scenarios, where preserving information requires the CSTs to collectively cover the channel set. Since $H_{\text{CST}}$ measures collective coverage rather than redundancy among individual tokens, we complement it with Gram-matrix redundancy and per-CST attention analyses in Sec.~\ref{sec:info_bottleneck_supp}, which directly diagnose whether memory tokens replicate the channel preferences of \texttt{[CLS]}. We report $H_{\texttt{CLS}}$ and $H_{\text{CST}}$ as percentages of the maximum entropy $\log C$, denoted $H_{\texttt{CLS}}\%$ and $H_{\text{CST}}\%$.

\noindent\textbf{Channel Dominance and Inter-Channel Redundancy.}

\begin{wrapfigure}[12]{r}{0.36\textwidth}
\centering
\captionsetup{font=small}
\vspace{-15pt}
\includegraphics[width=\linewidth]{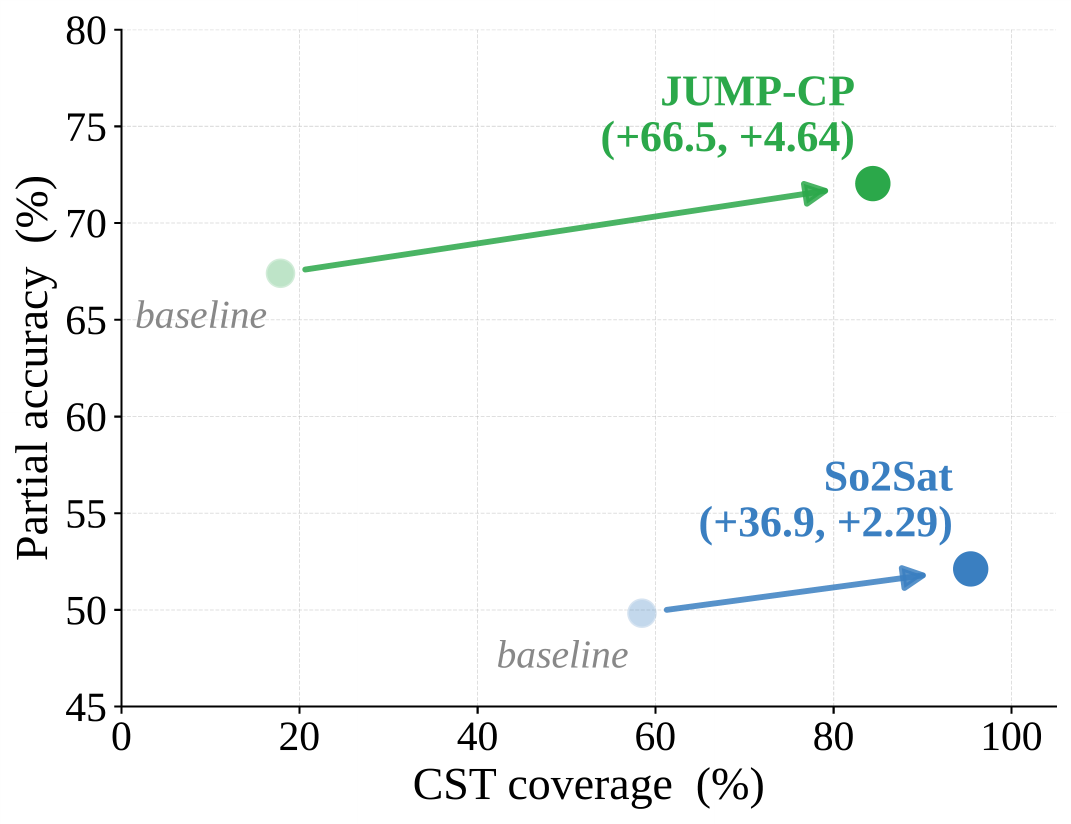}
\caption{
Restoring CST coverage ($\mathcal{H}_{\text{CST}}\%$, x-axis) drives partial-channel performance.
}
\label{fig:cst_coverage}
\end{wrapfigure}

Our analysis reveals that channel dominance emerges broadly in existing multi-channel ViTs (Fig.~\ref{fig:channel_collapse}(a)). On JUMP-CP and So2Sat, $H_{\texttt{CLS}}$ drops to $1.6\%$ and $6.4\%$, respectively, indicating that the global summary token is heavily concentrated on a small subset of channels rather than capturing diverse channel-wise information across the input. In contrast, CHAMMI ($C \in \{3,4\}$) exhibits substantially weaker collapse with $H_{\texttt{CLS}} = 98.3\%$, suggesting that a smaller channel set imposes a weaker structural bottleneck on the global token. This indicates that $H_{\texttt{CLS}}$ alone is insufficient to characterize multi-channel representation quality, since channel-specific information may remain entangled with shared structure even when token-level coverage appears intact.

Channel dominance is not merely a representational phenomenon but is also associated with downstream performance degradation. As shown in Fig. 1(b), mitigating collapse improves both entropy and classification accuracy on all three datasets, establishing a direct link between representation quality and downstream task performance. While $H_{\texttt{CLS}}$ diagnoses channel dominance at the representational level, whether the model truly captures distinct channel-specific information requires examining the underlying channel structure.

We next examine when channel dominance arises and what governs the extent of recoverable performance. Multi-channel data inherently couples globally shared structure and channel-specific information over the same spatial layout, which makes their separation a defining challenge of multi-channel representation learning. When inter-channel redundancy is high, channels carry largely overlapping content, so the model can rely on a small set of dominant channels as a shortcut, while fine-grained channel-specific cues are easily suppressed. Recovering these suppressed cues therefore requires not only mitigating channel dominance, but also a mechanism that interprets each channel as a distinct factor at the representation level, so that channel-specific information can be separated rather than absorbed into the shared structure. We quantify inter-channel redundancy $\rho$ as the off-diagonal mean of the per-image channel-wise cosine similarity matrix (formal definition in Appendix~\ref{app:channel_redundancy}). As shown in Fig.~\ref{fig:channel_collapse}(c), the magnitude of the performance gain scales with $\rho$, while remaining positive across the full redundancy spectrum spanned by three benchmarks ($\rho \in [0.19, 0.77]$).

We further examine whether memory tokens fulfill the role assumed in our earlier discussion, namely distributing channel-wise information across multiple tokens, as quantified by the collective coverage $\mathcal{H}_{\text{CST}}$ in Eq.~\eqref{eq:cst_coverage}. As shown in Fig.~\ref{fig:cst_coverage}, baseline CSTs exhibit $\mathcal{H}_{\text{CST}}$ substantially below the maximum on both JUMP-CP ($17.87\%$) and So2Sat ($58.48\%$), indicating that the CST set as a whole covers only a limited subset of the channel composition. This collective coverage deficit suggests that memory tokens do not specialize on distinct channels but instead inherit the dominance pattern of the \texttt{[CLS]} token. Sec.~\ref{sec:info_bottleneck_supp} further supports this view at the individual-token level on JUMP-CP, where both the \texttt{[CLS]} and memory tokens align with the same subset of dominant channels.

Taken together, these findings reveal a structural limitation of existing CSTs. The intended decomposition, where the [CLS] token captures globally shared semantics and memory tokens preserve diverse channel-wise cues, is not realized in practice. Instead, both token types collapse onto the same dominant channels, leaving channel-specific information entangled with the shared structure rather than disentangled into distinct channel factors. Effective multi-channel modeling therefore requires both explicit separation of channel information across CSTs and an adaptive mechanism that interprets each channel under the observed channel composition.

\vspace{-0.5em}
\begin{figure*}[t!]
  \centering
  \includegraphics[width=0.99\linewidth]{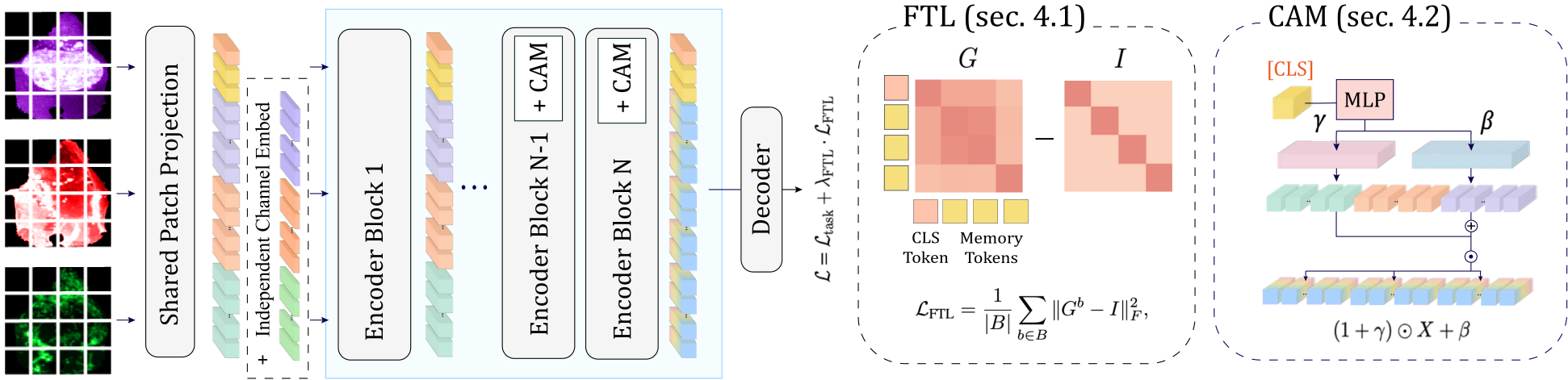}
  \caption{\textbf{Overall architecture of MuCa-ViT.} Multi-channel images are patchified using a shared projection and serialized with independent channel embeddings. Factorial Token Learning (FTL) enforces structured decorrelation among Context Semantic Tokens (CSTs), consisting of the global \texttt{[CLS]} token and memory tokens, via a token-wise Gram matrix constraint in \eqref{eq:loss_ftl}. Channel Composition-Aware Modulation (CAM) uses the FTL-refined \texttt{[CLS]} token as a semantic anchor to generate sample-wise modulation parameters $(\gamma, \beta)$, adaptively aligning patch representations under varying channel configurations.}
  \label{fig:architecture}
\end{figure*}

\section{MuCa-ViT: Multi-Channel Composition-Aware Vision Transformer}
\label{sec:method}

As analyzed in Sec.~\ref{sec:info_bottleneck}, multi-channel ViTs exhibit two coupled limitations: (i) context tokens collapse onto dominant channels, producing redundant CSTs that fail to capture distinct channel-specific information, and (ii) the model lacks a mechanism to interpret channel semantics under varying channel compositions, leaving channel-specific information entangled with the shared structure. These limitations prevent the model from maintaining diverse channel information and adapting representations to dynamic input configurations. To address them, we propose \textbf{MuCa-ViT}, a unified framework consisting of two complementary components: \textbf{Factorial Token Learning (FTL)} and \textbf{Channel Composition-Aware Modulation (CAM)}. FTL suppresses redundancy among CSTs to recover distinct channel-specific information across tokens, while CAM leverages this refined representation to disentangle channel semantics from the shared structure under the observed channel composition.

\vspace{-1em}
\subsection{Factorial Token Learning (FTL)}
\label{sec:method:ftl}
As discussed in Sec.~\ref{sec:info_bottleneck}, existing multi-channel ViTs suffer from channel dominance in which the \texttt{[CLS]} token and memory tokens align to a small subset of dominant channels, resulting in highly redundant Context Semantic Tokens (CSTs). This collapse undermines the role that CSTs would ideally play as a structured representation space, where global shared semantics and channel-specific information are distributed across multiple tokens. To address this limitation, we propose \textbf{Factorial Token Learning (FTL)},
which explicitly enforces token-level decorrelation by suppressing redundancy across the CST set. The key idea is to discourage different context tokens from encoding overlapping information, thereby encouraging them to organize complementary channel-specific semantic factors across CSTs. In this way, FTL restores the decomposition that CSTs would ideally support: the \texttt{[CLS]} token captures globally shared semantics, while memory tokens preserve complementary channel-specific cues.

To quantify redundancy among CSTs, we consider the set of context tokens $\mathrm{CST} \in \mathbb{R}^{(K+1) \times D}$, where $K$ is the number of memory tokens. We compute a token-wise Gram matrix $G_{ij} = \frac{1}{D} \langle \mathrm{CST}_i, \mathrm{CST}_j \rangle$, where large off-diagonal values indicate overlapping semantic information across tokens, and define a decorrelation objective:
\begin{equation}
\mathcal{L}_{\text{FTL}} = \frac{1}{|B|} \sum_{b\in B} \| W \odot (G^{b} - I) \|_F^2,
\qquad
W = \begin{bmatrix} 1 & \mathbf{1}^\top \\ \mathbf{1} & W' \end{bmatrix},\;\;\;\;\;
W'_{mn} = \begin{cases} 1, & m=n \\ \alpha, & m\neq n \end{cases}
\label{eq:loss_ftl}
\end{equation}
where $I$ is the identity matrix, $G^b$ is the Gram matrix of sample $b$ (batch-normalized beforehand), and $\alpha=0.1$ unless otherwise specified. This asymmetric weighting reflects the asymmetry in how task supervision propagates through the context tokens. The \texttt{[CLS]} token directly feeds into the prediction head and therefore receives the strongest task-loss gradient, which makes it most susceptible to absorbing dominant-channel signals as a shortcut. By enforcing decorrelation between \texttt{[CLS]} and memory tokens, FTL pushes channel-specific information away from \texttt{[CLS]} into the memory tokens, allowing \texttt{[CLS]} to attend more evenly across channels rather than collapsing onto the dominant ones. We thus prioritize \texttt{[CLS]}--memory decorrelation as the primary target of factorization, while treating memory--memory decorrelation as a secondary constraint to reduce local redundancy without suppressing useful cross-channel correlations, as validated in Appendix~\ref{sec:asym_ablation}. FTL is applied to the final encoder output and combined with the task loss
$\mathcal{L} = \mathcal{L}_{\mathrm{task}} + \lambda_{\mathrm{FTL}}\,\mathcal{L}_{\mathrm{FTL}}$.

By suppressing redundancy across the CST set, FTL mitigates the channel dominance identified in Sec.~\ref{sec:info_bottleneck} and recovers distinct
channel-specific information at the token level. However, distributing information across CSTs alone does not disentangle channel-specific information from the shared structure at the feature representation level. Achieving this further requires adapting the patch representations themselves to the observed channel composition, which we address with CAM in Sec.~\ref{sec:method:cam}.

\subsection{Channel Composition-Aware Modulation (CAM)}
\label{sec:method:cam}

While FTL recovers channel-specific information at the token level by suppressing redundancy across the CST set, it does not transform the patch representations themselves. The semantic role of each channel is inherently composition-dependent, and disentangling channel-specific information from the shared structure requires adapting the patch representations to the observed channel composition (Sec.~\ref{sec:info_bottleneck}).

Existing approaches rely on fixed channel embeddings, which treat channels independently of the input composition and fail to capture such contextual variations. In contrast, the \texttt{[CLS]} token aggregates globally shared semantics across varying channel configurations. However, its direct exposure to the task-loss gradient biases it toward dominant channels (Sec.~\ref{sec:info_bottleneck}), and only after FTL suppresses this bias does it serve as a reliable conditioning source. To this end, we propose \textbf{Channel Composition-Aware Modulation (CAM)}, which uses the FTL-refined \texttt{[CLS]} token as a global semantic anchor to interpret the input channel configuration and generate composition-aware, sample-adaptive modulation parameters for the patch representations.

For a multi-channel image, we denote the patch tokens of channel $c \in \{1,\dots,C\}$ as $\mathbf{X}_c \in \mathbb{R}^{P \times D}$, and stack them into $\mathbf{X} \in \mathbb{R}^{C \times P \times D}$. A common strategy~\cite{bao2023channel,Pham2024,pham2025cha} injects channel identity via channel embeddings, $\mathbf{Z}^{\mathrm{in}}_{patch} = \mathbf{X} + \mathbf{e}$, where $\mathbf{e} \in \mathbb{R}^{C \times D}$ is broadcast across patches. While this preserves channel identity, it does not capture how each channel should be interpreted under the observed channel composition.

To address this limitation, CAM generates modulation parameters from the \texttt{[CLS]} token at the $k^{th}$ encoder block:
\begin{equation}
(\gamma^k, \beta^k) = f_{\mathrm{CAM}}({\texttt{[CLS]}}^k),
\end{equation}
which are applied to all patch tokens. $f_{\mathrm{CAM}}(\cdot)$ denotes a lightweight conditioning network implemented as a two-layer MLP with a SiLU activation~\cite{ramachandran2017searching} in between.
Given the patch token set $\mathbf{Z}^k$, we perform feature-wise affine transformation:
\begin{equation}
    \mathbf{Z}^k_{patch} \leftarrow (1+\gamma^k)\odot \mathbf{Z}^k_{patch} + \beta^k,   \quad     \mathbf{Z}^k_{patch} \in \mathbb{R}^{(C\times P) \times D},
\end{equation}
where $\odot$ denotes element-wise multiplication.
In CAM, the modulation parameters $(\gamma, \beta)$ are generated in a sample-specific manner from the global context token \texttt{[CLS]}. This self-conditioned formulation allows the model to adapt feature transformations for each input based on its global semantics.
Importantly, since the \texttt{[CLS]} token reflects the active channel composition, the resulting modulation parameters are not only instance-specific but also composition-aware. We empirically validate this design choice in Appendix~\ref{sec:cond_ablation}.
By conditioning on this global representation, CAM adapts feature
transformations under each input configuration, enabling composition-aware channel interpretation that disentangles channel-specific information from the shared structure (Sec.~\ref{sec:spectral_analysis_disentangle}). As a result, the same channel can be interpreted differently under varying channel subsets.

\paragraph{Coupled Design of FTL and CAM.}
FTL and CAM are inherently coupled, operating at complementary levels. FTL recovers channel-specific information at the token level by mitigating channel dominance, while CAM transforms the patch representations themselves under the observed channel composition. Without FTL, the \texttt{[CLS]} token remains biased toward dominant channels, and CAM inherits this bias when generating its modulation parameters. Conversely, without CAM, recovering information across CSTs alone does not transform the patch representations to disentangle channel-specific information from the shared structure. Together, FTL and CAM address the two coupled limitations identified in Sec.~\ref{sec:info_bottleneck}, jointly enabling channel-specific information to be both distributed across CSTs and disentangled from the shared structure (Sec.~\ref{sec:spectral_analysis_disentangle}).
\section{Experiments}
\textbf{Datasets.}
We evaluated our method on three publicly available multi-channel benchmarks: \textbf{JUMP-CP}~\citep{chandrasekaran2024three}, an 8-channel microscopy dataset comprising five fluorescence and three brightfield channels, with 161 classes (160 perturbations and one control);
\textbf{CHAMMI}~\citep{chen2023chammi,viana2023integrated,thul2017subcellular,way2022morphology}, an integration of three microscopy datasets (3--5 channels) for domain generalization evaluated by the CHAMMI Score; and \textbf{So2Sat}~\citep{zhu2019so2sat}, an 18-channel SAR and multispectral remote sensing dataset with 17 climate zone classes.

\textbf{Implementation Details.}
We employ a ViT-Small backbone~\citep{dosovitskiy2020image} trained for 100 epochs on top of ChA-MAEViT~\citep{pham2025cha} and IC-ViT~\citep{lian2025isolated}, with $\lambda_{\mathrm{FTL}} = 0.1\mathrm{e}{-5}$ and CAM applied to the final Transformer block (last two blocks for JUMP-CP). Full configurations are provided in Appendix~\ref{app:experimental_details}.

\begin{table*}[t]
\centering
\small
\caption{{Comparison of test accuracy (\%) of channel-adaptive models.
\textbf{Full} denotes inference using all channels, while \textbf{Partial} evaluates on a subset of channels.
All results are reported as the mean $\pm$ standard deviation of three independent runs ($n=3$).
Values marked with $^*$ were reproduced using the official implementation.}}
\label{tab:channel_adaptive_results}
\setlength{\tabcolsep}{3pt}
\begin{tabular}{lccccc}
\toprule
\textbf{Model}
& \textbf{CHAMMI} ~\citep{chen2023chammi}
& \multicolumn{2}{c}{\textbf{JUMP-CP} ~\citep{chandrasekaran2024three}}
& \multicolumn{2}{c}{\textbf{So2Sat} ~\citep{zhu2019so2sat}} \\
\cmidrule(lr){3-4} \cmidrule(lr){5-6}
& \textbf{Avg score}
& \textbf{Full}
& \textbf{Partial}
& \textbf{Full}
& \textbf{Partial} \\
\midrule
HyperNetViT~\citep{ha2016hypernetworks}
& 56.08 $\pm$ 0.41
& 47.07 $\pm$ 0.47
& 42.43 $\pm$ 0.65
& 60.73 $\pm$ 0.24
& 41.88 $\pm$ 0.85 \\
DepthwiseViT~\citep{chen2023chammi}
& 61.80 $\pm$ 0.43
& 49.86 $\pm$ 0.45
& 44.98 $\pm$ 0.71
& 60.41 $\pm$ 0.22
& 43.41 $\pm$ 1.10 \\
TemplateMixingViT~\citep{pham2024mixturegrowth,plummer2020neural,savarese2019learning}
& 58.16 $\pm$ 0.42
& 52.48 $\pm$ 0.27
& 43.85 $\pm$ 0.73
& 55.86 $\pm$ 0.10
& 37.28 $\pm$ 0.34 \\
ChAda-ViT~\citep{bourriez2024chada}
& 63.93 $\pm$ 0.42
& 65.03 $\pm$ 0.98
& 42.15 $\pm$ 2.33
& 56.98 $\pm$ 0.46
& 12.38 $\pm$ 2.03 \\
ChannelViT~\citep{bao2023channel}
& 64.97 $\pm$ 0.58
& 67.51 $\pm$ 0.35
& 56.49 $\pm$ 0.53
& 61.03 $\pm$ 0.17
& 46.16 $\pm$ 0.40 \\
DiChaViT~\citep{Pham2024}
& 69.77 $\pm$ 0.44
& 69.19 $\pm$ 0.47
& 57.98 $\pm$ 0.41
& 63.36 $\pm$ 0.11
& 47.76 $\pm$ 0.23 \\
ChA-MAEViT~\citep{pham2025cha}
& 74.63 $\pm$ 0.54
& 90.73 $\pm$ 0.14
& 68.05 $\pm$ 0.21
& 67.44 $\pm$ 0.38
& 52.11 $\pm$ 0.49 \\
\midrule
ChA-MAEViT$^*$~\citep{pham2025cha}
& 73.95 $\pm$ 0.54
& 90.52 $\pm$ 0.40
& 67.40 $\pm$ 0.65
& 66.44 $\pm$ 0.74
& 49.83 $\pm$ 1.25 \\

\textbf{MuCa-ViT}
& \textbf{76.19 $\pm$ 0.37}
& \textbf{95.14 $\pm$ 0.32}
& \textbf{72.04 $\pm$ 0.14}
& \textbf{67.97 $\pm$ 0.68}
& \textbf{52.12 $\pm$ 0.95} \\
\bottomrule
\end{tabular}
\label{tab:main_results}
\end{table*}

\subsection{Main Results on Multi-Channel Benchmarks}
\label{sec:main_results}

A clear performance pattern emerges across datasets when compared to our reproduced ChA-MAEViT$^{*}$. Our analysis in Sec.~\ref{sec:info_bottleneck} identifies inter-channel redundancy $\rho$ (see Appendix~\ref{app:channel_redundancy}) as the factor governing how much channel-specific information has been suppressed and is therefore available for recovery. Consistent with this prediction, gains scale with $\rho$ across all three benchmarks (Fig.~\ref{fig:channel_collapse}(c)).

The three datasets each probe a distinct regime of our analysis. JUMP-CP ($\rho=0.77$, $H_{\texttt{CLS}}=1.6\%$) exhibits the most severe channel dominance and yields the largest gain, where both FTL and CAM operate on heavily collapsed representations.
CHAMMI ($\rho = 0.46$) presents a complementary case. Although its $H_{\texttt{CLS}}$ is already saturated, MuCa-ViT still yields a $+2.24$ pp improvement, demonstrating that channel-specific information remains entangled with the shared structure at the representation level even when token-level coverage is fully achieved. This regime, where token-level diagnostics alone cannot reveal the underlying entanglement, is precisely where representation-level disentanglement (Sec.~\ref{sec:spectral_analysis_disentangle}) becomes essential.
So2Sat ($\rho=0.19$) shows a smaller yet positive gain (+1.53 pp), confirming that the mechanism remains effective at the low-redundancy end. The same recovery is visible at the token level. MuCa-ViT raises $H_{\text{CST}}$ from $17.87\%$ to $84.41\%$ on JUMP-CP and from $58.48\%$ to $95.41\%$ on So2Sat (Fig.~\ref{fig:cst_coverage}), and the partial-channel gains in Table~\ref{tab:main_results} reflect the joint effect of this restored coverage and CAM's composition-aware adaptation.

\subsection{Ablation Study}
\begin{wraptable}{r}{0.5\textwidth}
\centering
\vspace{-1em}
\captionsetup{font=small}
\caption{Component ablation on JUMP-CP. CAM is applied to the final encoder block only.}
\label{tab:ablation}
\setlength{\tabcolsep}{8pt}
\small
\begin{tabular}{lcc}
\toprule
\textbf{Model} & \textbf{Full} & \textbf{Partial} \\
\midrule
ChA-MAEViT     & 90.35 & 68.10 \\
+ FTL          & 91.58 & 68.23 \\
+ CAM          & 92.46 & 70.96 \\
+ FTL + CAM    & \textbf{94.35} & \textbf{71.81} \\
\bottomrule
\end{tabular}
\vspace{-1em}
\end{wraptable}

\textbf{Ablation on FTL and CAM.}
Table~\ref{tab:ablation} evaluates the individual contributions of FTL and CAM on JUMP-CP. FTL alone provides moderate improvements through token-level decorrelation, addressing channel dominance across CSTs. CAM alone improves
performance by introducing composition-aware modulation at the representation level. However, the combination of FTL and CAM yields substantially higher performance than the sum of their individual gains, indicating that CAM benefits substantially from the conditioning signal refined by FTL.

This confirms that the two modules address the two coupled limitations identified in Sec.~\ref{sec:info_bottleneck} at complementary levels: FTL recovers channel-specific information at the token level, while CAM leverages this refined representation to disentangle channel-specific information from the shared structure at the feature representation level. Following the observation in CaiT~\citep{touvron2021cait} that the [CLS] token reaches semantic maturity in deeper blocks, we apply CAM to the later encoder blocks, where adaptation benefits more from composition-aware interpretation. We provide a detailed analysis of CAM's optimal insertion depth in Appendix~\ref{sec:block_ablation} and Appendix~\ref{sec:placement_ablation}.

\textbf{Validation of CAM.}
While CAM improves performance, we further verify that it performs composition-aware modulation rather than generic feature scaling. As shown in Appendix Fig.~\ref{fig:r_b_parameters}, the modulation parameters cluster according to channel configurations rather than sample identity, indicating that CAM adapts feature transformations based on channel composition.
This behavior provides direct evidence that CAM adapts feature transformations based on the underlying channel configuration. Together with the ablation results, this confirms that CAM functions as a composition-aware conditioning mechanism, complementing FTL in addressing the coupled limitations identified in Sec.~\ref{sec:info_bottleneck}.

\begin{figure}[t]
  \centering
  \includegraphics[width=0.99\linewidth]{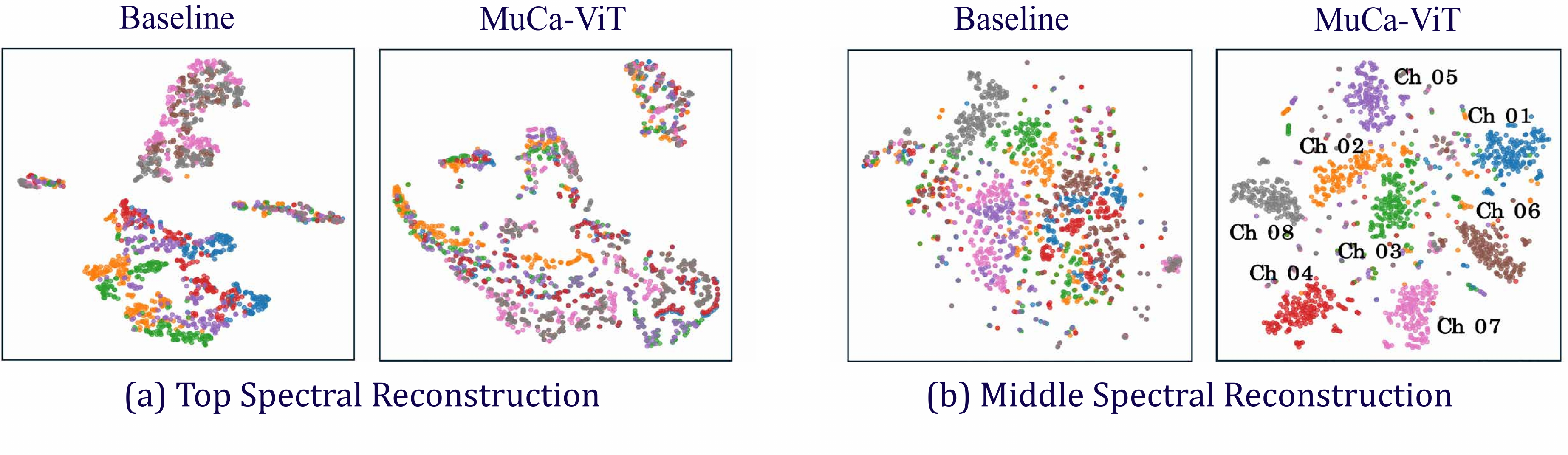}
  \vspace{-0.8em}
  \caption{SVD-based subspace analysis of multi-channel representations. The 384-dimensional singular spectrum is partitioned into 55 groups of 7 consecutive singular vectors each. \textbf{(a)} t-SNE of $\mathbf{Z}^N_{\mathrm{top}}$ (Group~1, $d_{\mathrm{top}}{=}7$), capturing globally shared channel semantics. \textbf{(b)} t-SNE of $\mathbf{Z}^N_{\mathrm{mid}}$ (Groups~3--35, $d_{\mathrm{mid1}}{=}15,\, d_{\mathrm{mid2}}{=}245$), corresponding to residual semantic components. In the middle spectral space, the baseline collapses into an entangled representation, whereas MuCa-ViT preserves clear channel-wise clustering.}
  \label{fig:CAM}
\end{figure}

\subsection{Spectral Analysis on Channel Disentanglement}
\label{sec:spectral_analysis_disentangle}

While Sec.~\ref{sec:info_bottleneck} analyzes channel dominance through token-level coverage, here we examine whether channel-specific information is also disentangled from the shared structure at the feature representation level. Multi-channel inputs couple globally shared structure and channel-specific semantics over the same spatial layout, and an effective multi-channel representation should separate these two factors rather than absorbing channel-specific information into the shared component. To examine how these factors are organized in the learned representation space, we
perform a spectral analysis based on singular value decomposition (SVD), with detailed formulation provided in
Appendix~\ref{sec:representation_analysis}.

As shown in Fig.~\ref{fig:CAM}, the top spectral component (a) exhibits substantial overlap across channels for both the baseline~\cite{pham2025cha} and our method, indicating that it primarily captures globally shared cross-channel structure. In contrast, the middle spectral component (b) reveals a clear difference: the baseline remains entangled across channels, whereas MuCa-ViT produces well-separated channel-wise clusters. This provides direct evidence that MuCa-ViT addresses the second limitation identified in Sec.~\ref{sec:info_bottleneck}, namely the entanglement of channel-specific information with the shared structure, while the top spectral component remains shared as expected.
This shared-private organization aligns with the factorization observed in multimodal representation learning~\cite{tsai2019multimodal,lai2024shared}, where shared and modality-specific factors are explicitly disentangled. Our results indicate that the coupled design of FTL and CAM induces this structure: FTL recovers channel-specific information across CSTs and refines the conditioning signal, which CAM then leverages to disentangle channel-specific information from the shared structure at the representation level.

\subsection{Plug-in Compatibility Across Architectures}

We evaluate the plug-in compatibility of FTL and CAM by integrating them into IC-ViT~\cite{lian2025isolated} and DiChaViT~\cite{Pham2024}. As shown in Table~\ref{tab:plugin}, our modules consistently improve performance across both architectures.
On JUMP-CP, incorporating FTL and CAM yields $+1.15$pp (full) and $+2.09$pp (partial) on IC-ViT$^{\text{mem}}$, and $+3.26$pp (full) and $+2.68$pp (partial) on DiChaViT$^{\text{mem}}$.
Here, IC-ViT$^{\text{mem}}$ and DiChaViT$^{\text{mem}}$ denote memory-token-augmented variants constructed to enable a controlled evaluation of FTL and CAM under matched architectural settings, as both modules operate on memory-token slots. Within this matched setting,
these results demonstrate that FTL and CAM are architecture-agnostic and can be seamlessly integrated into diverse multi-channel ViT models.

\begin{table}[t]
\centering
\caption{Plug-in compatibility of FTL and CAM on IC-ViT~\citep{lian2025isolated} and DiChaViT~\citep{Pham2024}. JUMP-CP is reported under full and partial channel settings. Models with superscript \textit{mem} denote architectures augmented with memory tokens using the same initialization as MuCa-ViT.}
\small
\label{tab:plugin}
\setlength{\tabcolsep}{16pt}
\begin{tabular}{lcc | lcc}
\toprule
\cmidrule(lr){1-3} \cmidrule(lr){4-6}
\textbf{Model} & \textbf{Full} & \textbf{Partial} & \textbf{Model} & \textbf{Full} & \textbf{Partial} \\
\midrule
IC-ViT$^{\text{mem}}$ & 86.68 & 66.41 & DiChaViT$^{\text{mem}}$ & 64.83 & 54.69 \\
+ FTL + CAM & \textbf{87.83} & \textbf{68.50} & + FTL + CAM & \textbf{68.09} & \textbf{57.37} \\
\bottomrule
\end{tabular}
\end{table}
\section{Conclusion}
We presented \textbf{MuCa-ViT}, a structured framework for multi-channel representation learning that addresses representation collapse and the lack of composition-aware channel interpretation in existing multi-channel ViTs. Our approach combines \textbf{Factorial Token Learning (FTL)}, which suppresses token-level redundancy to produce a reliable global representation, with \textbf{Channel Composition-Aware Modulation (CAM)}, which leverages this representation to adaptively interpret channel semantics under varying channel compositions. By conditioning feature transformations on the global context, CAM enables the same channel to be interpreted differently depending on the input configuration. Extensive experiments show that the gains are larger in higher-redundancy regimes, consistent with our analysis that inter-channel redundancy is a major structural factor that promotes CST collapse. Plug-in evaluations further demonstrate the compatibility of FTL and CAM with other multi-channel ViT architectures. Overall, our results suggest that effective multi-channel modeling benefits from both structured representation and composition-aware interpretation.

\begin{ack}
This research was supported by the G-LAMP Program of the National Research Foundation of Korea (NRF), funded by the Ministry of Education (RS-2025-25442252); the Bio \& Medical Technology Development Program of the NRF, funded by the Korean government (MSIT) (RS-2022-NR068424); the BK21 FOUR program of the NRF; and the AI Seoul Tech Research Support Program of the Seoul Future Foundation. This work was supported by the Korea Bio Data Station (K-BDS) with computing resources including technical support.
\end{ack}

{
\small
\bibliography{reference}
}

\clearpage
\newpage
\appendix
\section*{Appendix Overview}
\addcontentsline{toc}{section}{Appendix Overview}

This appendix complements the main paper as follows.

Sec.~\ref{app:experimental_details} details the experimental setup and reproduction protocol.

Sec.~\ref{sec:design_analysis} reports design analyses and ablations on FTL and CAM, covering memory-token count, asymmetric weighting, $\lambda_{\mathrm{FTL}}$ sensitivity, modulation parameters, conditioning source, and insertion depth.

Sec.~\ref{sec:extended_exp} provides additional results, including task-level CHAMMI scores, robustness across channel configurations, ViT-Base scaling, computational cost, and token-budget analysis.

Sec.~\ref{sec:representation_analysis} formalizes the SVD-based spectral analysis. It also evaluates missing-channel reconstruction and reinjection.

Sec.~\ref{app:channel_redundancy} quantifies inter-channel redundancy and its correlation with performance gains.

Sec.~\ref{sec:info_bottleneck_supp} presents token-level evidence of representation collapse via Gram-matrix and attention analyses. It also tests the robustness of channel coverage to similarity normalization and temperature, and evaluates prediction under final \texttt{[CLS]} corruption.

Sec.~\ref{sec:cam_algo} describes the CAM algorithm.

Sec.~\ref{app:limitation} discusses limitations and future work.

\section{Experimental Details}
\label{app:experimental_details}

\begin{table}[h]
\centering
\caption{Training and evaluation configurations used to reproduce ChA-MAEViT~\citep{pham2025cha} (denoted ChA-MAEViT$^*$ in Table~\ref{tab:main_results}). MuCa-ViT is trained under identical configurations except for the memory token count and the additional FTL/CAM hyperparameters specified in Appendix~\ref{sec:block_ablation},~\ref{sec:lambda_ablation}.}
\label{tab:training_config}
\small
\begin{tabular}{lccc}
\toprule
\textbf{Configuration} & \textbf{JUMP-CP} & \textbf{CHAMMI} & \textbf{So2Sat} \\
\midrule
\multicolumn{4}{l}{\textit{Hardware and batch}} \\
GPUs (NVIDIA A6000) & 2 & 1 & 1 \\
Train batch size & 32 & 80 & 128 \\
Eval batch size & 300 & 256 & 512 \\
Num workers & 1 & 3 & 3 \\
\midrule
\multicolumn{4}{l}{\textit{Optimization}} \\
Optimizer & AdamW & AdamW & AdamW \\
Learning rate & $4 \times 10^{-4}$ & $4 \times 10^{-4}$ & $4 \times 10^{-4}$ \\
LR warmup epochs & 10 & 0 & 0 \\
Max gradient norm & 3 & 1 & 1 \\
Training epochs & 100 & 100 & 100 \\
\midrule
\multicolumn{4}{l}{\textit{Architecture}} \\
Patch size & 16 & 16 & 8 \\
Encoder depth & 11 & 11 & 11 \\
Decoder depth & 1 & 1 & 1 \\
Memory tokens (baseline) & 4 & 4 & 8 \\
Memory tokens (MuCa-ViT) & 3 & 3 & 7 \\
\bottomrule
\end{tabular}
\end{table}

\textbf{Datasets.}
For JUMP-CP, the eight channels consist of five fluorescence and three brightfield channels. For So2Sat, we follow the city-split protocol with 352K training and 24K test samples.

\textbf{Memory Tokens Configuration.}
Following the setup described in Fig.~\ref{fig:memory_token_comparison}, we used 3, 3, and 7 memory tokens for JUMP-CP, CHAMMI, and So2Sat, respectively. We further applied orthogonal initialization to the memory tokens to reduce token correlation during optimization. For ChA-MAEViT~\citep{pham2025cha}, we follow the default memory token settings that yield the best performance, using 4, 4, and 8 tokens for JUMP-CP, CHAMMI, and So2Sat, respectively. For IC-ViT~\citep{lian2025isolated}, we use 4 memory tokens on the JUMP-CP dataset. Notably, MuCa-ViT used fewer memory tokens than ChA-MAEViT's original settings; thus, the gains reported in Table~\ref{tab:main_results} are not attributed to larger token budgets (further analyzed in Appendix~\ref{sec:memory_ablation}).

\textbf{CAM Insertion Depth and FTL Weighting.}
Across datasets, applying CAM to the final Transformer block yields the best overall performance. For JUMP-CP, however, applying CAM to the last two blocks (10--11) provides consistent additional gains; we therefore adopt this setting when reporting its results in Table~\ref{tab:main_results}. For architectures without memory tokens, we insert memory tokens with identical initialization and apply FTL and CAM in a plug-and-play manner, preserving the original training protocol.

\textbf{Downstream Tasks and Evaluation Protocol.}
Our method follows the training and evaluation settings of ChA-MAEViT~\citep{pham2025cha} and is implemented on top of the baseline. Across all datasets, we jointly optimize the reconstruction and classification objectives for 100 epochs; the same training length is used for the fine-tuning stage in the IC-ViT~\citep{lian2025isolated} plug-in experiments. Performance is measured by top-1 classification accuracy on JUMP-CP and So2Sat, and by macro-averaged F1 over the WTC-11 and HPA tasks for CHAMMI. We trained the models using the AdamW optimizer, minimizing cross-entropy loss for JUMP-CP and So2Sat, and proxy loss for CHAMMI, following the protocol of ChA-MAEViT~\cite{pham2025cha}.

\textbf{Hardware Setup and Reproduction Protocol.}
Experiments were conducted on a single NVIDIA A6000 GPU (CHAMMI, So2Sat) and dual A6000 GPUs (JUMP-CP). To ensure consistency, all $\Delta$ values reported in this paper are computed against our reproduced ChA-MAEViT~\citep{pham2025cha} (marked $\ast$ in Table~\ref{tab:main_results}) under the same training protocol. For completeness, Table~\ref{tab:main_results} also includes the published results, and MuCa-ViT outperforms ChA-MAEViT under both references on JUMP-CP and CHAMMI.

\section{Design Analysis}
\label{sec:design_analysis}

\subsection{Effect of the Number of Memory Tokens}
\label{sec:memory_ablation}
\begin{figure}[htbp]
  \centering
  \includegraphics[width=0.85\linewidth]{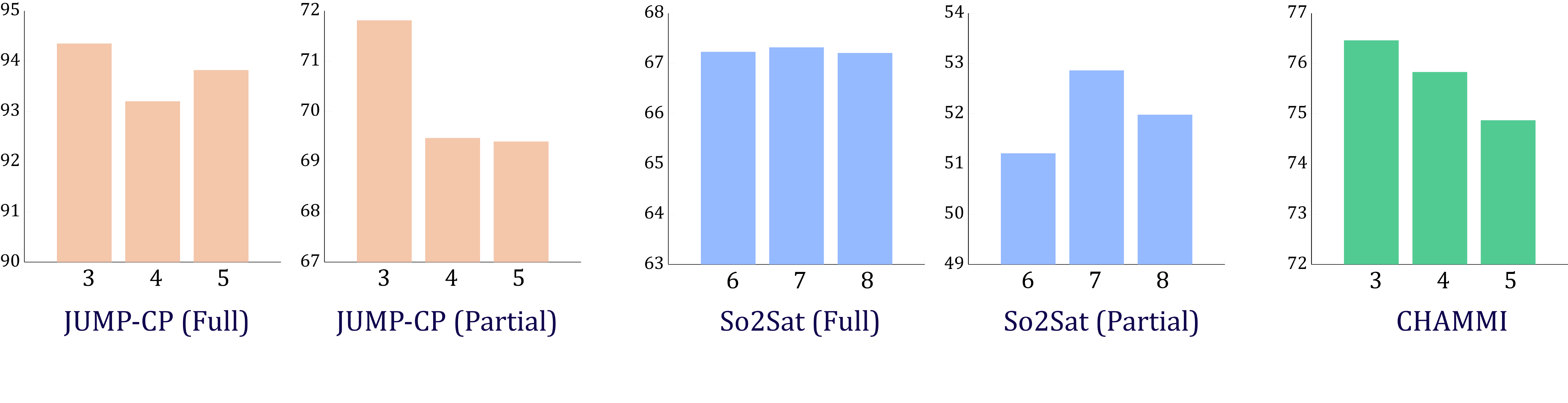}
  \vspace{-0.5cm}
  \caption{Effect of the number of memory tokens on JUMP-CP, So2Sat, and CHAMMI. Our model consistently achieves optimal performance with fewer memory tokens compared to the default settings of ChA-MAEViT.}
  \label{fig:memory_token_comparison}
\end{figure}

As shown in Fig.~\ref{fig:memory_token_comparison}, we analyze the effect of the number of memory tokens on the JUMP-CP, So2Sat, and CHAMMI datasets. In ChA-MAEViT~\cite{pham2025cha}, the default numbers of memory tokens are set to 4, 8, and 4 for JUMP-CP, So2Sat, and CHAMMI, respectively. In contrast, our model achieves the best performance with 3, 7, and 3 memory tokens on the corresponding datasets. This observation suggests that our method requires fewer memory tokens to achieve optimal performance. We attribute this to the reduced redundancy among CSTs enabled by FTL, which allows information to be more effectively separated and stored within each memory token. \daeun{Since the main comparison in Table~\ref{tab:main_results} uses each method's best-performing memory-token count, the comparison is favorable to the baseline: MuCa-ViT outperforms ChA-MAEViT despite using fewer memory tokens, so the reported gains cannot be attributed to a larger token budget.}

\subsection{Ablation Study on FTL}
\subsubsection{Ablation on Asymmetric Weight $\alpha$}
\label{sec:asym_ablation}
Analyzing the structure of inter-channel collapse reveals that [CLS]--memory decorrelation and memory--memory decorrelation serve distinct roles: [CLS]--memory decorrelation corrects channel bias in the global representation, while memory--memory decorrelation eliminates local redundancy between individual channel tokens. Treating $W$ (Eq.~\eqref{eq:loss_ftl}) as a uniform matrix with equal weights ignores this asymmetry. As shown in Table~\ref{tab:asymmetric}, the asymmetric weighting yields +1.32~pp / +0.45~pp over the equal-weight variant, validating that this design is not a cosmetic choice but a necessary consequence of the structure of inter-channel collapse.

\begin{table}[htbp]
\centering
\caption{Ablation on asymmetric weighting in FTL. JUMP-CP accuracy.}
\label{tab:asymmetric}
\small
\begin{tabular}{lcc}
\toprule
Variant & Full & Partial \\
\midrule
Equal-weight ablation      & 94.09 & 71.81 \\
\textbf{MuCa-ViT (asymmetric) }     & \textbf{95.41} & \textbf{72.26} \\
\bottomrule
\end{tabular}
\end{table}

\subsubsection{Sensitivity to Regularization Weight $\lambda_{FTL}$}
\label{sec:lambda_ablation}
\begin{figure}[htbp]
  \centering
  \includegraphics[width=0.4\linewidth]{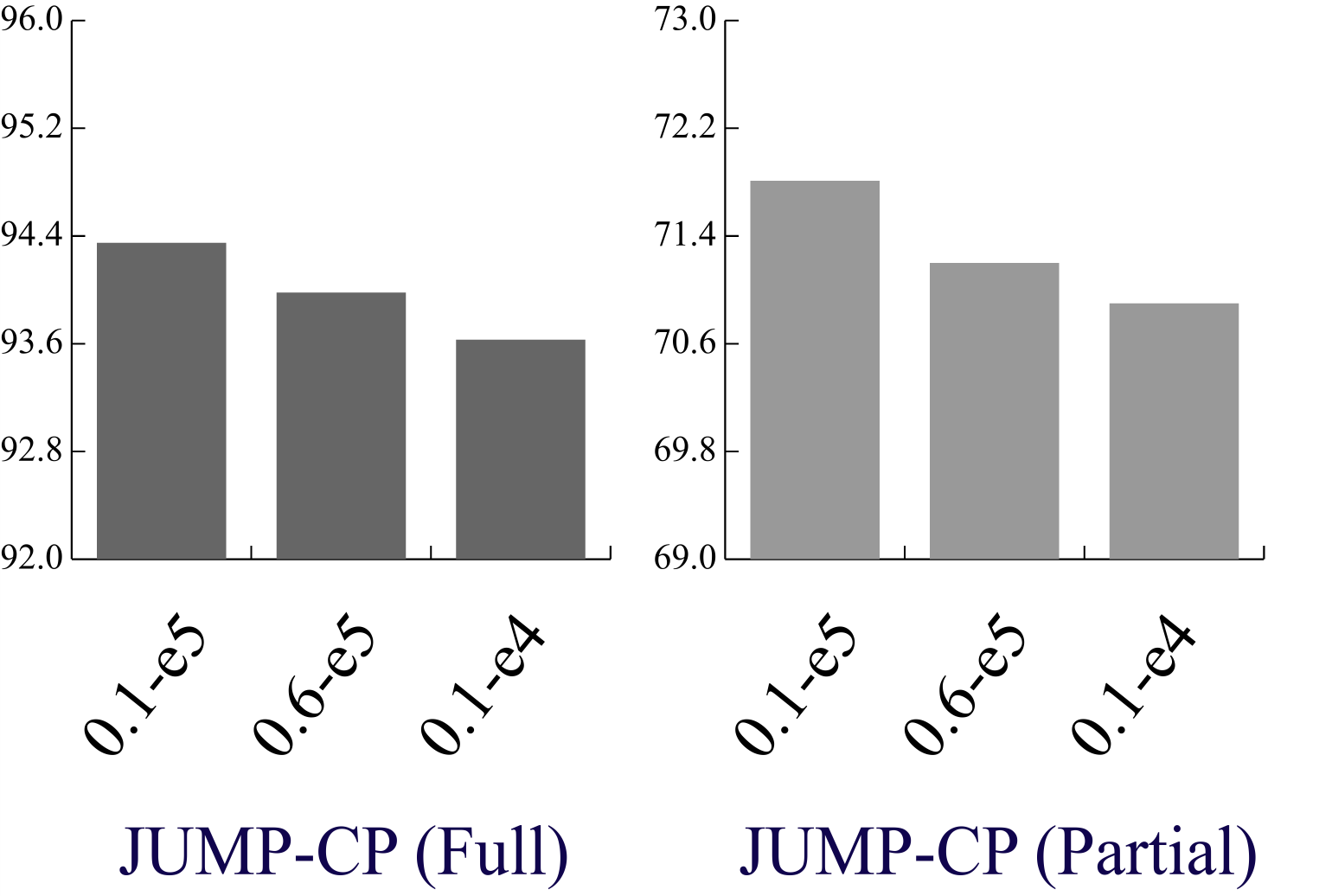}
  \caption{Sensitivity analysis of the FTL decorrelation weight $\lambda_{\mathrm{FTL}}$ on JUMP-CP under full and partial channel settings. Performance peaks at a moderate regularization strength and remains stable across a broad range of $\lambda_{\mathrm{FTL}}$.}
  \label{fig:ftl_loss_scale}
\end{figure}

We analyze the sensitivity of FTL to the decorrelation weight $\lambda_{\mathrm{FTL}}$ in Fig.~\ref{fig:ftl_loss_scale}. As shown in Fig.~\ref{fig:ftl_loss_scale}, performance peaks at a moderate regularization strength and gradually decreases as $\lambda_{\mathrm{FTL}}$ increases in both full- and partial-channel settings. This trend suggests that while structured decorrelation is beneficial for alleviating token redundancy, overly strong orthogonality constraints suppress useful task-relevant correlations. Importantly, the performance variation remains smooth and stable across a reasonable range of $\lambda_{\mathrm{FTL}}$, indicating that FTL does not rely on fragile hyperparameter tuning and achieves a balanced trade-off between semantic factorization and discriminative capacity.

\subsection{Ablation Study on CAM}
\subsubsection{Analysis of CAM Modulation Parameters}
\label{sec:cam_params}
\begin{figure}[htbp]
  \centering
  \includegraphics[width=0.9\linewidth]{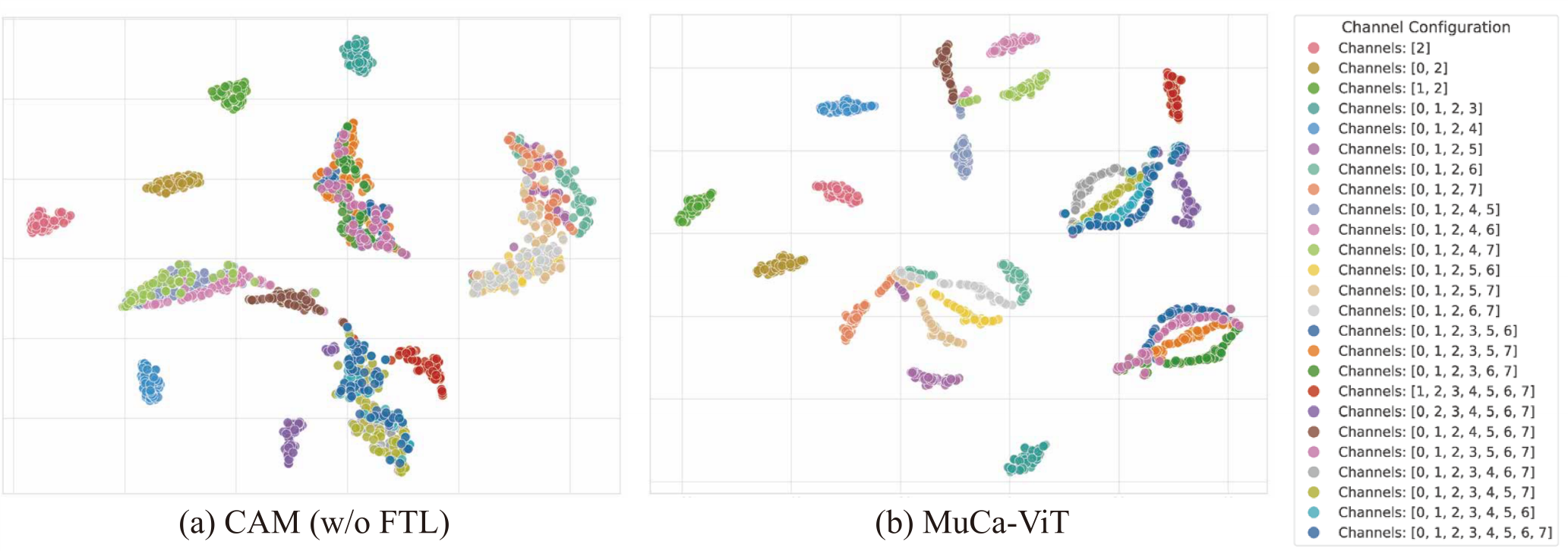}
  \caption{t-SNE visualization of the learned CAM modulation parameters ($\gamma, \beta$) across different channel configurations. \daeun{Each point is obtained by feeding the same underlying samples with a different channel subset.} Distinct clusters emerge for different channel subsets, \daeun{organized by configuration rather than sample identity}, indicating that CAM generates configuration-aware modulation signals rather than static embeddings.}
  \label{fig:r_b_parameters}
\end{figure}

To verify that CAM performs composition-dependent modulation, we visualize the learned affine parameters $(\gamma, \beta)$ using t-SNE in Fig.~\ref{fig:r_b_parameters}. Each point corresponds to the $(\gamma, \beta)$ generated by CAM when the \emph{same underlying samples} are fed with a \emph{different channel subset}; we vary the subset across all configurations shown in the legend and collect the resulting modulation parameters. As shown in Fig.~\ref{fig:r_b_parameters}, different channel subsets form clearly separable clusters, organized by channel configuration rather than by sample identity, indicating that CAM produces configuration-aware modulation signals rather than static adjustments. Configurations with overlapping channel sets exhibit related but distinct embeddings, suggesting that modulation parameters smoothly adapt to the global semantic context encoded in the \texttt{[CLS]} token. Comparing (a) CAM without FTL and (b) MuCa-ViT (CAM with FTL), the clusters become substantially sharper under FTL, confirming that the FTL-refined \texttt{[CLS]} provides a cleaner composition signal for modulation. These results confirm that CAM functions as an intrinsic self-conditioning mechanism, where the global semantic anchor dynamically governs feature recalibration under heterogeneous channel compositions.

\subsubsection{Ablation on Conditioning Source in CAM}
\label{sec:cond_ablation}

To isolate the contribution of FTL-refined self-conditioning, we compare three conditioning sources in Table~\ref{tab:cam_source}, organized along two axes. The first axis distinguishes \emph{external} conditioning, which uses signals provided alongside the input as explicit priors, from \emph{intrinsic} conditioning, which uses representations the model itself produces from the input, specifically the \texttt{[CLS]} token.

\paragraph{External vs.\ intrinsic conditioning.}
External conditioning approaches~\cite{perez2018film, dhariwal2021diffusion, peebles2023scalable, birnbaum2019temporal, zhang2025diffusion} typically rely on auxiliary inputs such as language embeddings, class labels, or timestep encodings as explicit priors. In our setup, we directly inject the channel composition as a multi-hot vector indicating the active channels, providing the model with explicit information about which channels are present. Intrinsic conditioning instead uses the model's own \texttt{[CLS]} token, which not only reflects the channel composition but also captures sample-specific semantics, allowing CAM to adapt its modulation per sample rather than relying on the channel composition alone. As shown in Table~\ref{tab:cam_source}, intrinsic conditioning with the FTL-refined \texttt{[CLS]} outperforms external conditioning by $+2.84$~pp, indicating that the global semantic representation built up within the model carries richer information about the channel configuration than an external categorical prior.

\paragraph{Effect of FTL refinement within intrinsic conditioning.}
Within intrinsic conditioning, we further compare two variants. The first uses the raw \texttt{[CLS]} token, which remains biased toward dominant channels, and the second uses the FTL-refined \texttt{[CLS]} token. The FTL-refined variant outperforms the raw \texttt{[CLS]} variant by $+2.23$~pp. This isolates the contribution of collapse correction within intrinsic conditioning and indicates that the effectiveness of CAM depends on the quality of the internal signal, not merely on whether the signal is internal.

\begin{table}[htbp]
\centering
\caption{Ablation on the conditioning source in CAM. JUMP-CP accuracy.}
\label{tab:cam_source}
\small
\begin{tabular}{lcc}
\toprule
Conditioning Source & Full & Partial \\
\midrule
External conditioning                              & 92.57 & 70.47 \\
Intrinsic conditioning (w/o FTL refinement)        & 93.18 & 71.12 \\
FTL-refined intrinsic conditioning (MuCa-ViT)      & \textbf{95.41} & \textbf{72.26} \\
\bottomrule
\end{tabular}
\end{table}

\subsubsection{Effect of CAM Insertion Depth in the Encoder}
\label{sec:block_ablation}
\begin{figure}[htbp]
  \centering
  \includegraphics[width=0.8\linewidth]{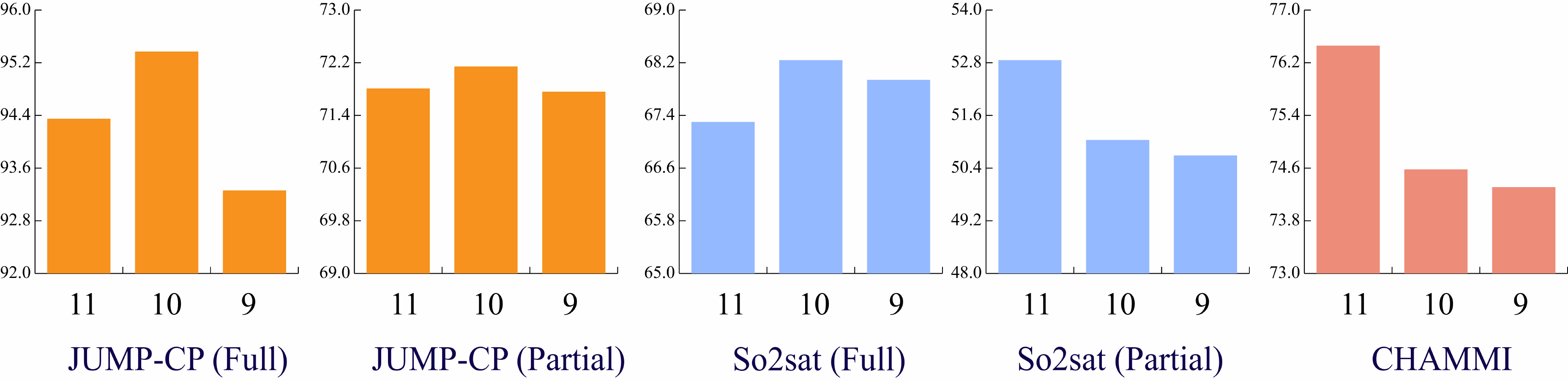}
  \caption{Effect of CAM insertion depth across different transformer blocks. Each x-axis label denotes the starting block from which CAM is applied to all subsequent blocks (e.g., Block 9 indicates CAM applied to blocks 9–11). }
  \label{fig:block_sensitivity}
\end{figure}

Fig.~\ref{fig:block_sensitivity} analyzes the impact of inserting CAM at different transformer blocks. As shown in Fig.~\ref{fig:block_sensitivity}, applying CAM to the final encoder block provides a strong default across datasets, while extending CAM to the last two blocks yields additional gains on JUMP-CP. Placing CAM too early limits its access to sufficiently abstract global semantics. These results support our design choice of inserting CAM near the final transformer stage, where the global \texttt{[CLS]} token encodes stable semantic context while still allowing effective feature recalibration.

We further observe that the optimal CAM \emph{placement depth} depends on inter-channel redundancy. Datasets with higher redundancy benefit from applying CAM to more blocks, while datasets with lower redundancy reach their best performance with CAM at the final block alone. Specifically, JUMP-CP ($\rho=0.77$, Table~\ref{tab:redundancy_summary}) benefits from extending CAM to the last two blocks, whereas CHAMMI ($\rho=0.46$) and So2Sat ($\rho=0.19$) achieve their best performance with CAM applied only to the final block.

\subsubsection{Ablation on CAM Placement across Network Depth}
\label{sec:placement_ablation}
Following the observation in CaiT~\cite{touvron2021cait} that the [CLS] token reaches its full semantic maturity in deeper blocks, we justify placing CAM in the last two blocks. Applying CAM in early layers can interfere with low-level feature extraction before the [CLS] token has accumulated sufficient semantic information to serve as a reliable conditioning signal. In contrast, placing CAM in deeper layers allows the model to leverage matured high-level abstractions for precise channel configuration. As shown in Table~\ref{tab:cam_depth}, bottom placement (last two blocks) yields a +3.23~pp / +1.60~pp improvement over top placement.

\begin{table}[htbp]
\centering
\caption{Ablation on CAM placement across network depth. JUMP-CP accuracy.}
\label{tab:cam_depth}
\small
\begin{tabular}{lcc}
\toprule
Variant & Full & Partial \\
\midrule
ChA-MAEViT (Baseline)                        & 90.35 & 68.10 \\
\;+ FTL (Ours w/o CAM)                       & 91.58 & 68.23 \\
\;+ FTL + CAM (Top, first 2 blocks)          & 92.18 & 70.66 \\
\;+ FTL + CAM (Bottom, last 2 blocks)        & \textbf{95.41} & \textbf{72.26} \\
\bottomrule
\end{tabular}
\end{table}

\section{Additional Results}
\label{sec:extended_exp}

\subsection{Detailed CHAMMI Results}
\label{sec:chammi_detailed}
In Table~\ref{tb:chammi_tasks}, we provide task-level F1 scores on the CHAMMI benchmark to complement the aggregated results reported in the main text. CHAMMI consists of multiple sub-datasets with varying channel compositions and domain shifts. Compared with our reproduced ChA-MAEViT$^{*}$, MuCa-ViT achieves higher scores on 6 of the 9 tasks, including all WTC and HPA tasks, and obtains the best result among all models on four of them.

\begin{table*}[htbp]
\centering
\caption{F1 scores of multi-channel ViT models on the CHAMMI benchmark.
ChA-MAEViT$^{*}$ denotes our reproduction under the same training environment as MuCa-ViT; both are single-run results.
\textbf{Bold} indicates the best result within each group, and \underline{underline} indicates the best result overall.}
\label{tb:chammi_tasks}
\setlength{\tabcolsep}{4pt}
\small
\begin{tabular}{lccccccccc}
\toprule
& \multicolumn{2}{c}{WTC~\citep{viana2023integrated}}
& \multicolumn{3}{c}{HPA~\citep{thul2017subcellular}}
& \multicolumn{4}{c}{CP~\citep{way2022morphology}} \\
\cmidrule(lr){2-3} \cmidrule(lr){4-6} \cmidrule(lr){7-10}
\textbf{Model}
& Task1 & Task2
& Task1 & Task2 & Task3
& Task1 & Task2 & Task3 & Task4 \\
\midrule
HyperNetViT~\citep{ha2016hypernetworks}       & 58.83 & 45.78 & 88.78 & 82.70 & 52.52 & 82.13 & 53.74 & 23.16 & 7.42 \\
DepthwiseViT~\citep{chen2023chammi}      & 69.81 & 52.19 & 91.65 & \textbf{\underline{88.04}} & 54.78 & 81.24 & 54.08 & \textbf{\underline{23.21}} & 5.87 \\
TemplateMixingViT~\citep{pham2024mixturegrowth,plummer2020neural,savarese2019learning}     & 61.66 & 51.52 & 85.01 & 79.91 & 49.69 & 77.45 & 48.83 & 22.56 & 5.60 \\
ChAda-ViT~\citep{bourriez2024chada}         & 77.58 & 67.18 & 87.49 & 75.94 & 45.41 & 83.92 & 45.58 & 21.94 & 6.28 \\
ChannelViT~\citep{bao2023channel}        & 78.36 & 67.58 & 83.93 & 76.73 & 47.97 & 77.70 & 55.16 & 21.89 & 6.38 \\
DiChaViT~\citep{Pham2024}          & 80.87 & 75.18 & 88.08 & 79.26 & 49.45 & 84.08 & 53.03 & 20.95 & 5.60 \\
ChA-MAEViT~\citep{pham2025cha}        & \textbf{84.52} & \textbf{77.15} & \textbf{94.14} & 87.47 & \textbf{56.75} & \textbf{90.89} & \textbf{\underline{56.68}} & 10.25 & \textbf{7.50} \\
\midrule
ChA-MAEViT$^{*}$~\citep{pham2025cha}  & 83.42 & 76.00 & 96.12 & 86.94 & 57.20 & \textbf{\underline{96.89}} & \textbf{48.19} & 10.12 & \textbf{\underline{8.90}} \\
MuCa-ViT (Ours)   & \textbf{\underline{85.98}} & \textbf{\underline{80.29}} & \textbf{\underline{96.27}} & \textbf{87.95} & \textbf{\underline{57.34}} & 96.59 & 45.30 & \textbf{11.15} & 8.51 \\
\bottomrule
\end{tabular}
\end{table*}

\subsection{Evaluation under Various Channel Configurations}
\label{sec:various_configs}
To evaluate robustness under missing-channel scenarios, we evaluated the model on all channel configurations derived from the full set of eight channels on JUMP-CP. We vary the number of available channels and consider all corresponding channel combinations, reporting the mean and standard deviation across configurations in Table~\ref{tb:diff_ch}. As shown in Table~\ref{tb:diff_ch}, MuCa-ViT consistently outperforms the baseline across nearly all channel budgets, with the largest gains observed in intermediate regimes (5--7 channels) where channel composition variability is most challenging.

\begin{table*}[htbp]
\centering
\caption{Performance comparison across different numbers and combinations of channels at inference. Mean $\pm$ std over all channel combinations at each budget.}
\label{tb:diff_ch}
\setlength{\tabcolsep}{1pt}
\small
\begin{tabular}{lcccccccc}
\toprule
& \multicolumn{8}{c}{\textbf{Number of channels at inference}} \\
\cmidrule(lr){2-9}
\textbf{Method} & \textbf{8} & \textbf{7} & \textbf{6} & \textbf{5} & \textbf{4} & \textbf{3} & \textbf{2} & \textbf{1} \\
\midrule
DiChaViT~\citep{Pham2024}          & 69.19 & 61.91 {\tiny$\pm$9.3}  & 54.49 {\tiny$\pm$12.4} & 46.35 {\tiny$\pm$13.4} & 38.00 {\tiny$\pm$12.4} & 30.09 {\tiny$\pm$9.3}  & 23.97 {\tiny$\pm$4.9} & 20.90 {\tiny$\pm$1.6} \\
ChA-MAEViT~\citep{pham2025cha}        & 90.73 & 83.36 {\tiny$\pm$8.3}  & 74.55 {\tiny$\pm$11.7} & 63.46 {\tiny$\pm$13.8} & 50.85 {\tiny$\pm$13.9} & 38.13 {\tiny$\pm$11.2} & 27.62 {\tiny$\pm$6.4} & \textbf{21.59} {\tiny$\pm$2.1} \\
\midrule
\textbf{MuCa-ViT} & \textbf{95.41} & \textbf{89.46} {\tiny$\pm$6.28} & \textbf{80.06} {\tiny$\pm$11.32} & \textbf{69.75} {\tiny$\pm$12.84} & \textbf{55.84} {\tiny$\pm$13.80} & \textbf{41.18} {\tiny$\pm$11.65} & \textbf{28.79} {\tiny$\pm$6.84} & 21.54 {\tiny$\pm$2.25} \\
\bottomrule
\end{tabular}
\end{table*}

\subsection{Scaling to ViT-Base}
\label{sec:vit_base}

We further evaluate our method at the ViT-Base scale, with results summarized in Table~\ref{tab:vit_base}. ChA-MAEViT$^*$ results in this section are reproduced by us under the same training environment as MuCa-ViT. MuCa-ViT consistently improves over ChA-MAEViT$^*$ at the ViT-Base scale on both JUMP-CP and CHAMMI, indicating that the gains from FTL and CAM are not specific to a particular backbone capacity.

\begin{table}[htbp]
\centering
\caption{Performance at ViT-Base scale.}
\label{tab:vit_base}
\small
\begin{tabular}{lccc}
\toprule
Model & JUMP-CP (Full) & JUMP-CP (Partial) & CHAMMI \\
\midrule
ChA-MAEViT*       & 92.22 & 70.24 & 76.20 \\
\textbf{MuCa-ViT (Ours)}  & \textbf{94.28} & \textbf{74.31} & \textbf{76.72} \\
\bottomrule
\end{tabular}
\end{table}

\subsection{Computational Complexity and Inference Latency}
\label{sec:complexity}
We evaluate the computational overhead and inference latency at batch size $32$ on ViT-Small, with results reported in Table~\ref{tab:complexity}. The CAM modules introduce an additional latency of approximately $0.15$s per batch, while the overall FLOPs remain effectively unchanged. We provide a detailed token-budget analysis in Appendix~\ref{sec:cost_budget}, which shows
that the gains are not attributable to additional capacity.

\begin{table}[htbp]
\centering
\caption{Computational complexity and inference latency at ViT-Small (batch size 32).}
\label{tab:complexity}
\small
\setlength{\tabcolsep}{15pt}
\begin{tabular}{lccc}
\toprule
Model & Inference Latency & FLOPs \\
\midrule
ChA-MAEViT       & 1.084s & 249.48 \\
\textbf{MuCa-ViT (Ours)} & 1.235s & \textbf{249.33} \\
\bottomrule
\end{tabular}
\end{table}

\subsection{Computational Cost and Token Budget Analysis}
\label{sec:cost_budget}

\begin{table}[htbp]
\centering
\caption{Performance comparison under matched and best-performing memory-token budgets. MuCa-ViT outperforms ChA-MAEViT$^*$ at the same token budget and exceeds its best-performing configuration with fewer memory tokens. ChA-MAEViT$^*$ denotes our reproduced baseline. Results are from single runs.}
\label{tb:mem_matched}
\setlength{\tabcolsep}{12pt}
\small
\begin{tabular}{lcccc}
\toprule
\multirow{2}{*}{\textbf{Model}} & \multirow{2}{*}{\textbf{mem}}
& \multicolumn{2}{c}{\textbf{JUMP-CP}} & \multirow{2}{*}{\textbf{CHAMMI}} \\
\cmidrule(lr){3-4}
& & Full & Partial & \\
\midrule
ChA-MAEViT$^*$           & 3 & 90.24 & 67.35 & 72.66 \\
ChA-MAEViT$^*$ (best)    & 4 & 90.35 & 68.10 & 74.04 \\
\textbf{MuCa-ViT (Ours)} & 3 & \textbf{95.41} & \textbf{72.26} & \textbf{76.46} \\
\bottomrule
\end{tabular}
\end{table}

\paragraph{Information density, not capacity, drives the gains.}
ChA-MAEViT$^*$ requires 4 memory tokens to reach its best configuration, and reducing the budget from 4 to 3 produces a consistent drop ($-0.11$/$-0.75$ pp on JUMP-CP Full/Partial, and $-1.38$ pp on CHAMMI), confirming that its representation quality depends on additional token slots. MuCa-ViT exhibits the opposite behavior. It reaches its best performance with only 3 memory tokens and still outperforms the best-configured ChA-MAEViT$^*$ by +5.06/+4.16 pp on JUMP-CP and +2.42 pp on CHAMMI. Since MuCa-ViT improves under a reduced token budget where the baseline degrades, the observed gains cannot be attributed to additional capacity. They reflect a higher information density per token, induced by FTL's suppression of redundancy among Context Semantic Tokens (Sec.~\ref{sec:info_bottleneck}). Once each memory token is encouraged to encode information complementary to \texttt{[CLS]} rather than remaining redundant with it, fewer tokens suffice to cover the channel set.

\paragraph{Computational cost.}
The overall FLOPs remain effectively unchanged from the baseline ($249.33$ vs.~$249.48$~G, a $0.06\%$ difference), even though MuCa-ViT introduces the CAM modules. This is because the reduction by one memory token slightly shortens the self-attention sequence and offsets the additional cost of CAM. Latency increases by approximately $0.15$s per batch at batch size $32$, reflecting the sequential overhead of inserting CAM at the late transformer blocks.

\section{Representation Analysis}
\label{sec:representation_analysis}
This section provides the formal details of the spectral analysis introduced in Sec.~\ref{sec:spectral_analysis_disentangle}, including the SVD-based decomposition formulation and the quantitative diversity metric used in Fig.~\ref{fig:CAM}.

Multi-channel visual representations contain two fundamentally different
sources of variation: globally shared structural information common across
channels, and heterogeneous channel-specific semantics. To examine how these
factors are encoded and separated in the learned representation space, we
conduct a spectral analysis based on singular value decomposition
(SVD), which has been widely adopted to probe semantic structure in
representation learning~\cite{zhang2025assessing,yun2025soma}.

\paragraph{Spectral decomposition.}
Let the final patch output at the $N^{th}$ encoder block be
$\mathbf{Z} \in \mathbb{R}^{(C \times P) \times D}$, where patch tokens
from all channels are concatenated along the token dimension. We compute
the SVD as
\begin{equation}
    \mathbf{Z} = U \Sigma V^\top,
\end{equation}
where the right singular vectors $V \in \mathbb{R}^{D \times D}$ span the
principal semantic directions of the representation space. Following prior
spectral grouping strategies, we partition this basis into dominant and
residual components,
\begin{equation}
    V_{\mathrm{top}} = V[:,\, 1\!:\!d_{\mathrm{top}}], \quad
    V_{\mathrm{mid}} = V[:,\, d_{\mathrm{mid1}}\!:\!d_{\mathrm{mid2}}],
    \label{eq:top_mid}
\end{equation}
where $d_{\mathrm{top}}$ denotes the number of dominant singular directions
capturing globally shared structure, and $[d_{\mathrm{mid1}},
d_{\mathrm{mid2}}]$ specifies the residual spectral band used to analyze
channel-specific semantic variation. The reconstructed spectral component
for a group $g \in \{\mathrm{top}, \mathrm{mid}\}$ is given by
\begin{equation}
    \widetilde{\mathbf{Z}}_{g} = \left(\mathbf{Z} - \mu\right) V_g V_g^\top + \mu,
\end{equation}
where $\mu$ is the token-wise mean and $V_g \in \mathbb{R}^{D \times d_g}$
contains the selected right singular vectors. We use these reconstructions
to probe shared and channel-specific structure both qualitatively
(Sec.~\ref{sec:spectral_visualization}) and quantitatively
(Sec.~\ref{sec:spectral_diversity}).

\subsection{Visual Separation of Shared and Channel-Specific Semantics}
\label{sec:spectral_visualization}
We first visualize the reconstructed components $\widetilde{\mathbf{Z}}_{\mathrm{top}}$
and $\widetilde{\mathbf{Z}}_{\mathrm{mid}}$ via t-SNE in
Fig.~\ref{fig:CAM}. As shown in Fig.~\ref{fig:CAM}~(a), the top spectral
component exhibits substantial overlap across channels for both the
baseline~\cite{pham2025cha} and our method, indicating that it primarily
encodes globally shared cross-channel structure. In contrast,
Fig.~\ref{fig:CAM}~(b) shows that the middle spectral component yields
clear channel-wise clustering under our method, whereas the baseline
exhibits substantially weaker separation. This contrast indicates that
while shared structure is consistently captured by the dominant spectrum,
our method isolates heterogeneous channel-specific semantics within the
residual spectral subspace.

\subsection{Quantitative Spectral Diversity}
\label{sec:spectral_diversity}

To quantify this separation across spectral groups, we measure the
inter-channel diversity of each reconstructed component. Letting
$\widetilde{\mathbf{Z}}_{g,c} \in \mathbb{R}^{P \times D}$ denote the
tokens belonging to channel $c$ within $\widetilde{\mathbf{Z}}_{g}$, we
compute the mean patch token
\begin{equation}
    m_{g,c} = \frac{1}{P} \sum \widetilde{\mathbf{Z}}_{g,c},
\end{equation}
and define the inter-channel similarity and diversity as
\begin{equation}
    \mathrm{Sim}(g) = \frac{1}{C(C-1)} \sum_{c \neq c'} \cos\!\left(m_{g,c}, m_{g,c'}\right),
    \quad \mathrm{Div}(g) = 1 - \mathrm{Sim}(g).
\end{equation}

\begin{figure}[htbp]
    \centering
    \includegraphics[width=0.40\linewidth]{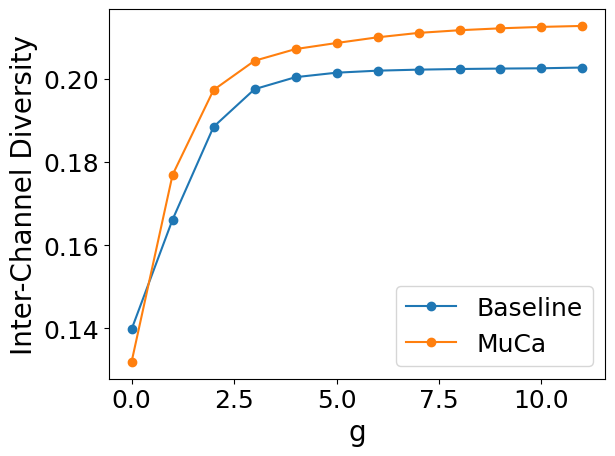}
    \caption{Inter-channel diversity $\mathrm{Div}(g)$ of reconstructed
    spectral components across spectral groups $g$ on JUMP-CP. MuCa-ViT
    (orange) maintains consistently higher diversity than the baseline
    (blue), especially in the middle spectral bands.}
    \label{fig:spectral_diversity}
\end{figure}

As shown in Fig.~\ref{fig:spectral_diversity}, the diversity curve reveals
a structural transition in the latent subspace. In the \textbf{early
spectral bands}, the reconstructed components exhibit high similarity
across channels, primarily encoding shared spatial structure. As we move
toward the \textbf{middle spectral bands}, a \textbf{distinct increase} in
diversity emerges for MuCa-ViT, while the baseline saturates at a
substantially lower level. This confirms quantitatively what the t-SNE
visualization shows the per-channel reconstructions become increasingly
distinct under our method, as MuCa-ViT captures discriminative
channel-specific semantics within these residual bands.

\paragraph{Connection to multimodal factorization.}
This behavior aligns with the shared–private factorization commonly
observed in multimodal representation learning~\cite{lai2024shared}.
In our setting, the dominant spectrum corresponds to semantics shared
across channels, whereas the middle spectrum captures fine-grained,
channel-specific variation. Notably, such spectral disentanglement emerges
only under our method and correlates with improved downstream performance,
suggesting that FTL and CAM jointly promote semantically meaningful
factorization in the representation space.

\subsection{Missing-Channel Reconstruction and Reinjection}
\label{app:reconstruction_reinjection}

The spectral analysis above examines how shared and channel-specific information is organized in the representation space. We further evaluate whether the learned memory representations preserve complementary information that can recover missing channels and support downstream prediction. Following the reconstruction protocol of ChA-MAEViT~\citep{pham2025cha}, we provide only the three brightfield channels of JUMP-CP and reconstruct the five missing fluorescence channels using the learned memory tokens. At test time, the reconstructed fluorescence channels are concatenated with the observed brightfield channels and reinjected into the frozen classifier.

\begin{table}[htbp]
\centering
\caption{Missing-channel reconstruction and reinjection on JUMP-CP. Five fluorescence channels are reconstructed from three observed brightfield channels and reinjected into the frozen classifier. Reinjection gain is measured relative to brightfield-only accuracy.}
\label{tab:reconstruction_reinjection}
\setlength{\tabcolsep}{10pt}
\small
\begin{tabular}{lccccc}
\toprule
\textbf{Method}
& \shortstack{\textbf{Brightfield-only}\\\textbf{Acc. (\%)}}
& \shortstack{\textbf{After reinjection}\\\textbf{Acc. (\%)}}
& \shortstack{\textbf{Gain}\\\textbf{(pp)}}
& \textbf{MSE} $\downarrow$
& \shortstack{\textbf{Fourier}\\\textbf{error} $\downarrow$} \\
\midrule
ChA-MAEViT
& 22.77 & 34.12 & +11.35 & 0.028 & 0.343 \\
\textbf{MuCa-ViT (Ours)}
& \textbf{23.54} & \textbf{37.39} & \textbf{+13.85}
& \textbf{0.027} & \textbf{0.321} \\
\bottomrule
\end{tabular}
\end{table}

As shown in Table~\ref{tab:reconstruction_reinjection}, MuCa-ViT obtains a larger reinjection gain (+13.85~pp) than ChA-MAEViT (+11.35~pp), while reducing both spatial reconstruction error and Fourier error. This improvement indicates that the memory representations preserve complementary information useful for recovering the missing fluorescence channels. Together with the spectral organization in Sec.~\ref{sec:spectral_visualization}, these results support the functional value of preserving channel-specific information alongside shared structural cues.

\section{Channel Redundancy Analysis}
\label{app:channel_redundancy}
This section provides the concrete measurement procedure and per-dataset values for the inter-channel redundancy $\rho$ used in Sec.~\ref{sec:info_bottleneck}.

A central observation of this work is that, when the input channels carry substantial overlapping information, training a multi-channel ViT without an explicit decorrelation mechanism is associated with \texttt{[CLS]} token collapse, as quantified by our entropy-based proxies in Sec.~\ref{sec:info_bottleneck}. To support this observation, we need a way to quantify how much redundancy each dataset actually contains. To this end, we measure channel-channel cosine similarity directly on the input images, using $16{\times}16$ patches.

For each dataset, we take normalized images $x \in \mathbb{R}^{C \times H \times W}$ and partition each image into $N$ non-overlapping $16 \times 16$ patches. For each image, we compute the per-image channel-channel similarity matrix $S \in \mathbb{R}^{C \times C}$ as
\begin{equation}
S_{ij} = \frac{1}{N} \sum_{p=1}^{N} \cos\!\bigl(x_i^p, x_j^p\bigr), \qquad i, j \in \{1, \dots, C\},
\end{equation}
where $x_i^p \in \mathbb{R}^{256}$ is the flattened $16 \times 16$ patch of channel $i$ at location $p$, and $\cos(u, v) = \langle u, v \rangle / (\|u\| \, \|v\|)$. Intuitively, at each patch location we compare the channel-$i$ and channel-$j$ patches, and then average over all $N$ patch locations. We then average $S$ over $2{,}048$ randomly sampled images per dataset to obtain $\bar{S} \in \mathbb{R}^{C \times C}$, and summarize $\bar{S}$ by its off-diagonal mean
\begin{equation}
\mathrm{Redundancy} = \frac{1}{C(C-1)} \sum_{i \neq j} \bar{S}_{ij},
\end{equation}
which captures the average similarity between distinct channels.

\begin{figure}[ht]
\centering
\includegraphics[width=0.7\linewidth]{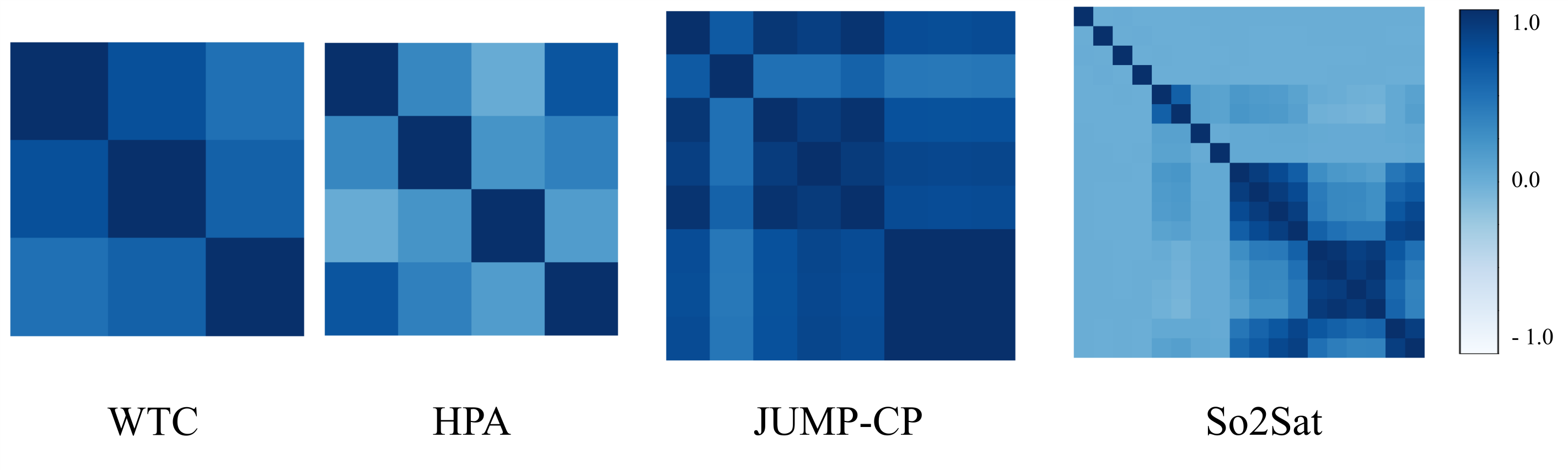}
\caption{Per-image channel-channel cosine similarity averaged over $2{,}048$ samples for each evaluation group. Darker blue indicates higher similarity. JUMP-CP and CHAMMI/WTC exhibit strong redundancy across channels, whereas So2Sat is largely uncorrelated except for a few pairs}
\label{fig:channel_cossim}
\end{figure}

\paragraph{Per-dataset cosine similarity matrices.}
Figure~\ref{fig:channel_cossim} shows the resulting matrices for the four evaluation groups, with CHAMMI's WTC and HPA chunks shown separately and JUMP-CP and So2Sat shown at the dataset level. Darker blue indicates higher cosine similarity between two channels. The structural differences across datasets are striking. JUMP-CP appears almost uniformly dark, indicating that its eight channels share substantial information. CHAMMI/WTC also exhibits strong redundancy, CHAMMI/HPA is mixed, and So2Sat is dominated by near-zero off-diagonal entries, with only a small number of strongly similar pairs (e.g., within the dual-polarization Sentinel-1 channels).

\paragraph{Quantitative summary.}
Table~\ref{tab:redundancy_summary} reports the off-diagonal mean similarity together with the number of channels and the absolute performance gain of MuCa-ViT over our reproduced ChA-MAEViT* baseline on the Full evaluation setting from Table~\ref{tab:main_results}. Across the three benchmarks, the average cosine similarity ranges from $0.19$ (So2Sat) to $0.77$ (JUMP-CP). For CHAMMI, we report the average over its WTC and HPA chunks and visualized separately in Fig.~\ref{fig:channel_cossim}.

\begin{table}[ht]
\centering
\caption{Channel redundancy and corresponding performance gain. \textbf{Redundancy} is the off-diagonal mean of the channel-channel cosine similarity matrix (Fig.~\ref{fig:channel_cossim}). $\Delta$\textbf{Full Acc.} is computed against our reproduced ChA-MAEViT* under the same training environment (cf.\ Table~\ref{tab:main_results}).}
\label{tab:redundancy_summary}
\setlength{\tabcolsep}{10pt}
\small
\begin{tabular}{lcccc}
\toprule
\textbf{Dataset / chunk} & \textbf{$C$} & \textbf{Redundancy} & \textbf{ChA-MAEViT*} & \textbf{$\Delta$Full Acc.}\\
\midrule
So2Sat~\citep{zhu2019so2sat}        & 18  & 0.19 & 66.44 & $+1.53$ \\
CHAMMI (avg)~\citep{{chen2023chammi}}  & 3-4 & 0.46 & 73.95 & $+2.24$ \\
JUMP-CP~\citep{chandrasekaran2024three}      &  8  & 0.77 & 90.52 & $+4.62$ \\
\bottomrule
\end{tabular}
\end{table}

\paragraph{Redundancy aligns with performance gain.}
Table~\ref{tab:redundancy_summary} reports the relationship between channel redundancy and the improvement $\Delta$Full Acc. across the three benchmarks. Datasets with higher channel redundancy show substantially larger gains. JUMP-CP, with the highest redundancy ($0.77$), benefits the most ($+4.62$). CHAMMI lies in the middle ($+2.24$ at $0.46$), and So2Sat, whose 18 channels carry largely orthogonal information, shows
a positive improvement ($+1.53$) at the low-redundancy end of the range.
The three benchmarks span a wide redundancy range from $0.19$ to $0.77$, and the gain ordering follows the redundancy ordering.

\section{Information Bottleneck in Context Semantic Tokens}
\label{sec:info_bottleneck_supp}

This section provides supplementary evidence supporting the CST collapse phenomenon analyzed in Sec.~\ref{sec:info_bottleneck}.

Multi-channel images substantially increase the length of the patch
sequence as the number of channels grows, forcing increasingly rich
semantic information to be summarized by a limited set of context tokens.
In ViT-based architectures~\cite{dosovitskiy2020image}, the \texttt{[CLS]} token
serves as a global semantic abstraction for downstream tasks. However,
when shared spatial semantics and channel-specific semantics are
entangled within a long token sequence, a single \texttt{[CLS]} token
inevitably becomes an information bottleneck and is insufficient to
summarize heterogeneous multi-channel information.

\begin{figure}[t]
    \centering
    \includegraphics[width=0.6\linewidth]{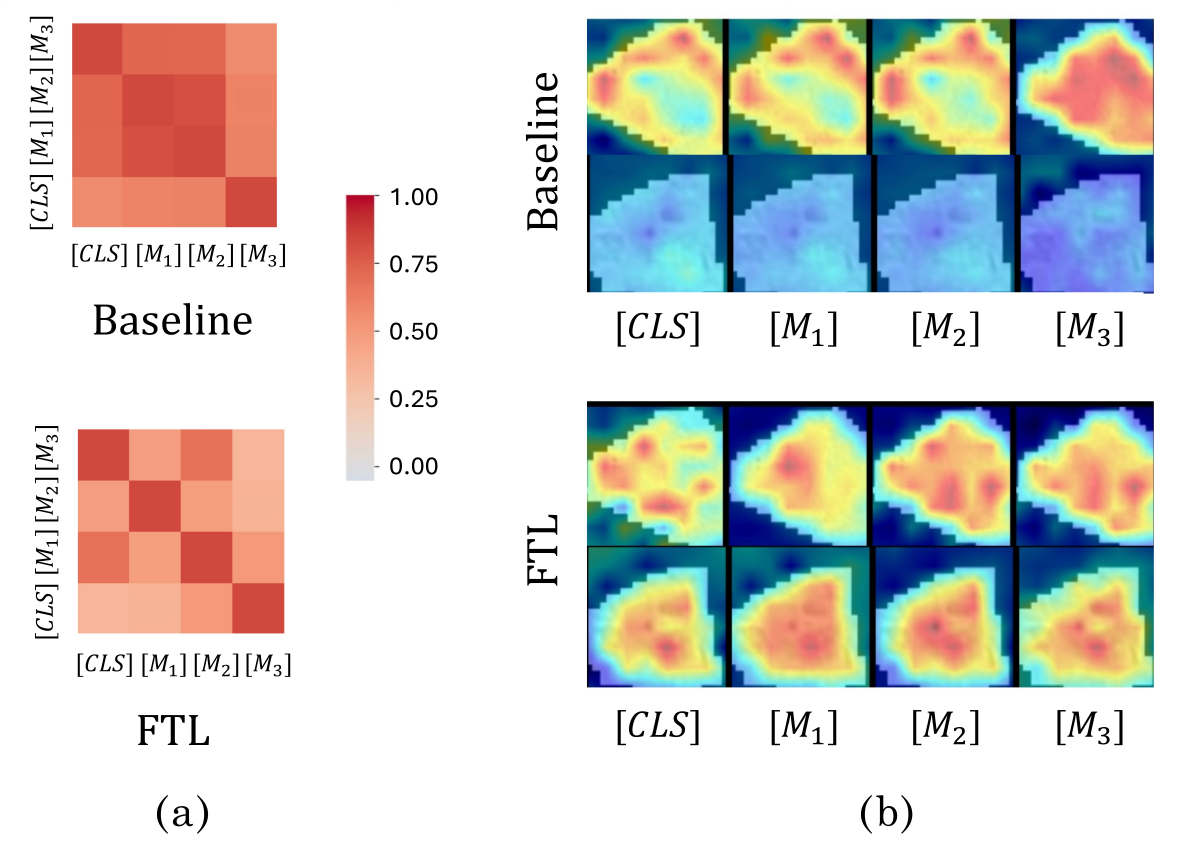}
    \vspace{-0.3cm}
    \caption{Comparison of Context Semantic Token (CST) behaviors
    between baseline ChA-MAEViT~\cite{pham2025cha} and our method.
    \textbf{(a)} Gram-matrix-based redundancy analysis among CSTs.
    \textbf{(b)} Attention heatmaps showing spatial focus for each token.}
    \label{fig:redundancy}
\end{figure}

Prior work introduces memory tokens to improve robustness under
missing-channel conditions~\cite{pham2025cha}. Here, we reinterpret
memory tokens as semantic storage slots intended to alleviate the token
bottleneck by distributing multi-channel information across multiple
context tokens. Nevertheless, in the absence of explicit structural
constraints, these memory tokens become highly redundant with the
\texttt{[CLS]} token, as shown in the baseline of
Fig.~\ref{fig:redundancy}~(a), resulting in low information density and failing to relieve the information bottleneck.

\begin{figure*}[t]
    \centering
    \includegraphics[width=1.0\linewidth]{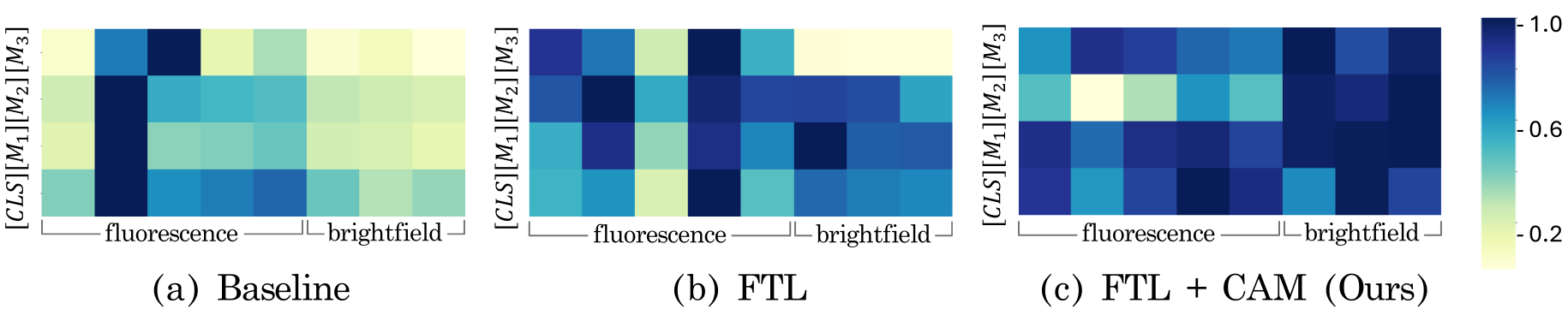}
    \vspace{-0.5cm}
    \caption{Dot-product similarity between channel-wise aggregated patch
    representations (x-axis) and the Context Semantic Tokens, including
    the \texttt{[CLS]} token and memory tokens $M_i$ $(i=1,2,3)$. In the
    baseline, both \texttt{[CLS]} and memory tokens exhibit strong
    alignment toward dominant channels, indicating severe semantic
    collapse and redundancy, whereas the proposed method yields
    well-disentangled similarity distributions.}
    \label{fig:dotproduct}
\end{figure*}

We analyze this redundancy from two complementary perspectives. First,
channel-wise attention heatmap analysis reveals that Context Semantic
Tokens consistently attend to the same dominant spatial regions rather
than capturing distinct semantic factors across channels in the baseline
of Fig.~\ref{fig:redundancy}~(b), indicating a lack of functional
specialization and semantic role separation. Second, from an information
storage perspective, we measure the dot-product similarity between CSTs
and channel-wise aggregated patch representations, and observe that
both the \texttt{[CLS]} token and memory tokens collapse toward a small
set of dominant channels in the baseline
(Fig.~\ref{fig:dotproduct}). Only a subset of channel tokens contributes
to semantic understanding, exhibiting severe semantic collapse.

Taken together, naive expansion of CSTs fails to distribute multi-channel semantics and instead biases the representation toward dominant channels. These observations suggest that effective multi-channel representation learning requires explicit structural constraints that suppress
redundancy and promote semantic factorization, with redundancy suppressed across CSTs so that memory tokens are encouraged to encode information complementary to \texttt{[CLS]}.

\subsection{Robustness of Channel Coverage to Similarity Normalization and Temperature}
\label{app:coverage_temperature}

The coverage measures in Eqs.~\eqref{eq:sim}--\eqref{eq:cst_coverage} use dot-product similarities, which depend on both directional alignment and representation norms. To examine whether the observed coverage advantage persists after removing norm effects, we recompute $H_{\texttt{CLS}}$ and $H_{\mathrm{CST}}$ using cosine-normalized similarities at multiple softmax temperatures. Specifically, we replace the patch-level distribution in Eq.~\eqref{eq:sim} with
\begin{equation}
\alpha_{c,p}^{\cos,\tau}(\mathbf{q}) =
\frac{\exp\!\left(\cos(\mathbf{q},\mathbf{Z}_{c,p})/\tau\right)}
{\sum_{c',p'}\exp\!\left(\cos(\mathbf{q},\mathbf{Z}_{c',p'})/\tau\right)},
\qquad \tau\in\{0.01,0.02,0.04,0.06\},
\label{eq:cosine_coverage_logits}
\end{equation}
and retain the channel aggregation and entropy definitions in Eqs.~\eqref{eq:sim}--\eqref{eq:cst_coverage}.

\begin{table}[htbp]
\centering
\caption{Cosine-based channel coverage across softmax temperatures. Coverage is reported as a percentage of maximum entropy, and $\Delta$ denotes the improvement of MuCa-ViT over ChA-MAEViT in percentage points. MuCa-ViT maintains higher coverage at every tested temperature.}
\label{tab:coverage_temperature}
\setlength{\tabcolsep}{10pt}
\small
\begin{tabular}{ccccccc}
\toprule
& \multicolumn{3}{c}{$H_{\texttt{CLS}}\%$}
& \multicolumn{3}{c}{$H_{\mathrm{CST}}\%$} \\
\cmidrule(lr){2-4}\cmidrule(lr){5-7}
$\tau$
& ChA-MAEViT & MuCa-ViT & $\Delta$
& ChA-MAEViT & MuCa-ViT & $\Delta$ \\
\midrule
\multicolumn{7}{l}{\textbf{JUMP-CP}} \\
0.01 & 10.23 & \textbf{42.31} & +32.08
     & 39.74 & \textbf{76.65} & +36.91 \\
0.02 & 28.79 & \textbf{63.96} & +35.17
     & 57.07 & \textbf{89.10} & +32.03 \\
0.04 & 60.66 & \textbf{88.74} & +28.08
     & 78.64 & \textbf{98.54} & +19.90 \\
0.06 & 81.93 & \textbf{96.61} & +14.68
     & 90.17 & \textbf{99.63} & +9.46 \\
\midrule
\multicolumn{7}{l}{\textbf{So2Sat}} \\
0.01 & 40.30 & \textbf{42.67} & +2.37
     & 70.83 & \textbf{82.92} & +12.09 \\
0.02 & 69.05 & \textbf{71.56} & +2.51
     & 82.26 & \textbf{93.42} & +11.16 \\
0.04 & 88.43 & \textbf{90.58} & +2.15
     & 91.41 & \textbf{98.33} & +6.92 \\
0.06 & 94.09 & \textbf{95.56} & +1.47
     & 94.97 & \textbf{99.34} & +4.37 \\
\bottomrule
\end{tabular}
\end{table}

Table~\ref{tab:coverage_temperature} shows that MuCa-ViT achieves higher $H_{\texttt{CLS}}$ and $H_{\mathrm{CST}}$ on both datasets at every tested temperature. Although the absolute entropy increases as the softmax distribution becomes less concentrated, the ordering between the two models remains unchanged. The cosine-based analysis therefore complements the dot-product diagnostic in Sec.~\ref{sec:info_bottleneck}, showing that the broader channel coverage is not explained solely by representation magnitude.

\subsection{Prediction under Final \texttt{[CLS]} Corruption}
\label{app:cls_noise}

The token-level analyses above examine how information is distributed across CSTs. We next test whether the intact memory-token pathway supports prediction when the final global representation becomes unreliable. After the encoder's final LayerNorm, we replace only the \texttt{[CLS]} output with
\begin{equation}
\widetilde{\mathbf{z}}_{\mathrm{CLS}} =
\sqrt{1-\eta}\,\mathbf{z}_{\mathrm{CLS}} +
\sqrt{\eta}\,\boldsymbol{\epsilon},
\qquad \boldsymbol{\epsilon}\sim\mathcal{N}(\mathbf{0},I),
\quad \eta\in\{0.5,0.7\}.
\label{eq:cls_corruption}
\end{equation}
All final memory-token outputs are kept intact. The corrupted token is then used as the query in the prediction-time pooling module,
\begin{equation}
\mathbf{z}_{\mathrm{pool}} =
\mathrm{CrossAttn}\!\left(
q=\widetilde{\mathbf{z}}_{\mathrm{CLS}},
K=[\mathbf{M}_1,\ldots,\mathbf{M}_K],
V=[\mathbf{M}_1,\ldots,\mathbf{M}_K]
\right),
\label{eq:cls_corruption_pooling}
\end{equation}
and the pooled representation is passed to the classifier. We evaluate both Full and Partial settings on JUMP-CP.

\begin{table}[htbp]
\centering
\caption{Prediction under final \texttt{[CLS]} corruption on JUMP-CP. Accuracy is reported in percent; the gap and parenthesized changes from clean accuracy are reported in percentage points.}
\label{tab:cls_corruption}
\setlength{\tabcolsep}{8pt}
\small
\begin{tabular}{llccc}
\toprule
\textbf{Condition}
& \textbf{Setting}
& \textbf{ChA-MAEViT}
& \textbf{MuCa-ViT}
& \textbf{Gap} \\
\midrule
\multirow{2}{*}{Clean}
& Full    & 90.35 & \textbf{95.41} & +5.06 \\
& Partial & 68.10 & \textbf{72.26} & +4.16 \\
\midrule
\multirow{2}{*}{$\eta=0.5$}
& Full    & 83.24 ($-7.11$) & \textbf{89.69} ($-5.72$) & +6.45 \\
& Partial & 58.67 ($-9.43$) & \textbf{67.78} ($-4.48$) & +9.11 \\
\midrule
\multirow{2}{*}{$\eta=0.7$}
& Full    & 71.64 ($-18.71$) & \textbf{80.14} ($-15.27$) & +8.50 \\
& Partial & 48.68 ($-19.42$) & \textbf{58.06} ($-14.20$) & +9.38 \\
\bottomrule
\end{tabular}
\end{table}

As shown in Table~\ref{tab:cls_corruption}, MuCa-ViT exhibits a smaller accuracy drop at both corruption levels. The difference is particularly pronounced under Partial evaluation, where the performance gap increases from 4.16~pp in the clean setting to 9.11~pp at $\eta=0.5$ and 9.38~pp at $\eta=0.7$. These results provide functional evidence that the memory-token pathway preserves information complementary to the final \texttt{[CLS]} representation, allowing prediction to remain more reliable when the global token is perturbed.

\section{Algorithmic Details of CAM}
\label{sec:cam_algo}
We implement CAM as a lightweight feature-wise modulation block inserted after the self-attention layer in selected transformer blocks; the full procedure is summarized in Algorithm~\ref{alg:cam_block}. Given the semantic context tokens $x$, we extract the global \texttt{[CLS]} token to generate modulation parameters $(\gamma, \beta)$, which are then applied to patch tokens through affine transformation.

CAM first projects the \texttt{[CLS]} token through a lightweight network to generate feature-wise modulation parameters:
\begin{equation}
(\gamma, \beta) = f_{\mathrm{CAM}}(\texttt{[CLS]}), \quad \gamma, \beta \in \mathbb{R}^{D}.
\end{equation}
$f_{\mathrm{CAM}}$ consists of a linear projection to dimension $D$, followed by a SiLU activation~\cite{ramachandran2017searching}, and a final linear layer outputting scaling and bias terms $(\gamma, \beta)$. These parameters are then broadcast and applied to each patch token via an affine transformation (see lines 4--5 in Algorithm~\ref{alg:cam_block}).

Unlike external conditioning approaches~\cite{perez2018film,
dhariwal2021diffusion, peebles2023scalable} that require auxiliary modules such as text encoders or label embedding networks to provide the conditioning signal, CAM derives $(\gamma, \beta)$ directly from the \texttt{[CLS]} token, enabling instance-adaptive channel interpretation without any additional encoder. The effectiveness of this internal signal depends on the quality of the \texttt{[CLS]} representation. FTL refines \texttt{[CLS]} into a reliable summary of cross-channel shared semantics by suppressing dominant-channel bias under high inter-channel redundancy (Sec.~\ref{sec:info_bottleneck}). The empirical importance of this refinement is verified in Appendix~\ref{sec:cond_ablation}.

As discussed in Sec.~\ref{sec:info_bottleneck}, the raw \texttt{[CLS]} token is biased toward dominant channels under high inter-channel redundancy. FTL is therefore introduced to refine \texttt{[CLS]} into a reliable summary of cross-channel shared semantics, which then serves as the conditioning signal for CAM. The empirical importance of this refinement is verified in Appendix~\ref{sec:cond_ablation}, where FTL-refined conditioning outperforms both raw \texttt{[CLS]} conditioning and external multi-hot conditioning.

\begin{algorithm}[htbp]
\caption{Channel Composition-Aware Modulation (CAM)}
\label{alg:cam_block}
\begin{algorithmic}[1]
\Require Token set $\mathbf{Z} \in \mathbb{R}^{(1+K+CP)\times D}$
\Ensure Modulated token set $\mathbf{Z}' \in \mathbb{R}^{(1+K+CP)\times D}$

\Statex \textbf{Notation:}
\Statex $\odot$: element-wise multiplication
\Statex $(\gamma, \beta) \in \mathbb{R}^{D}$ are broadcast over all $(C \times P)$ patch tokens along the feature dimension.
\Statex $\mathrm{Linear}$: a linear layer.
\Statex $C$, $P$, $D$: the number of channels, the number of patches per channel, the embedding dimension
\Statex $K$: the number of memory tokens.

\Function{$\mathrm{CAM}$}{$\mathbf{Z}$}
    \State $\texttt{[CLS]} \gets \mathbf{Z}[0]$
    \State $\mathbf{X} \gets \mathbf{Z}[1+K:1+CP+K]$
    \State $(\gamma, \beta) \gets \mathrm{Linear}(\mathrm{SiLU}(\mathrm{Linear}(\texttt{[CLS]})))$
    \State $\mathbf{X} \gets (1+\gamma) \odot \mathbf{X} + \beta$
    \State \Return $\mathbf{Z}' \gets \mathrm{Concat}(\texttt{[CLS]}, \mathbf{Z}[1\!:\!1+K], \mathbf{X})$
\EndFunction
\vspace{4pt}
\State $\mathbf{Y} \gets \mathrm{Attn}(\mathrm{Norm}_1(\mathbf{Z}), \text{mask})$
\State $\mathbf{Z} \gets \mathbf{Z} + \mathrm{DropPath}(\mathrm{LS}_1(\mathbf{Y}))$
\State $\mathbf{Z} \gets \mathrm{CAM}(\mathbf{Z})$
\State $\mathbf{Z} \gets \mathrm{Norm}_2(\mathbf{Z})$
\State $\mathbf{Z} \gets \mathbf{Z} + \mathrm{DropPath}(\mathrm{LS}_2(\mathrm{Linear}(\mathbf{Z})))$
\end{algorithmic}
\end{algorithm}

\section{Limitations and Future Work}
\label{app:limitation}
Our framework opens several directions for extension.

\textbf{First}, the underlying mechanism of token-level redundancy under high inter-channel redundancy and composition-dependent channel interpretation is not specific to fluorescence microscopy or remote sensing. Domains such as hyperspectral imaging, multi-parametric MRI, and multiplexed tissue imaging exhibit similar structural properties and represent natural targets for MuCa-ViT-style constraints, particularly as channel counts grow beyond what a single \texttt{[CLS]} token can summarize.

\textbf{Second}, the FTL hyperparameters are shared across JUMP-CP, CHAMMI, and So2Sat. A common final-block CAM configuration provides consistent gains across datasets, while extending CAM to the last two blocks is an additional refinement for JUMP-CP. Making the CAM insertion depth learnable, for example through block-wise gating, would enable the model to adapt its modulation structure without manually selecting the insertion depth.

\end{document}